\documentclass{article} %
\usepackage{iclr2023_conference,times}

\usepackage{hyperref}
\usepackage{url}

\newif\ifformatarxiv
\formatarxivtrue

\ifformatarxiv
    \iclrfinalcopy %
\fi

\usepackage{amsfonts}
\usepackage{amsmath}
\usepackage{pifont}%

\usepackage{xspace}

\usepackage{framed,enumitem} 

\usepackage{amssymb}%
\usepackage{pifont}%
\newcommand{\cmark}{\ding{51}}%
\newcommand{\xmark}{\ding{55}}%

\usepackage{caption}
\usepackage{subcaption}

\usepackage{algorithm}
\usepackage[noend]{algpseudocode}

\usepackage{wrapfig}
\usepackage{graphicx}
\usepackage{sistyle}
\SIthousandsep{,}

\usepackage{enumitem}
\setlist[itemize]{leftmargin=20pt}

\newcommand{\tldr}{\textbf{\underline{TL;DR}:~}}

\newcommand\pegasussmall{$\text{PEGASUS}_\text{SMALL}$\xspace}
\newcommand\pegasusbase{$\text{PEGASUS}_\text{BASE}$\xspace}
\newcommand\pegasuslarge{$\text{PEGASUS}_\text{LARGE}$\xspace}
\newcommand\pegasusxxlarge{$\text{PEGASUS}_{\text{2B}}$\xspace}

\newcommand\smallmodelsize{50M\xspace}
\newcommand\basemodelsize{200M\xspace}
\newcommand\largemodelsize{500M\xspace}
\newcommand\xxlargemodelsize{2B\xspace}

\newcommand\bartmodelsize{340M\xspace}

\newcommand\stmoemodelsize{268B\xspace}
\newcommand\ullmodelsize{20B\xspace}
\newcommand\unilmmodelsize{110M\xspace}

\newcommand\cnn{CNN/DailyMail\xspace}
\newcommand\xsum{XSUM\xspace}
\newcommand\reddit{RedditTIFU-long\xspace}
\newcommand\samsum{SAMSum\xspace}

\newcommand\qg{SQuAD QG\xspace}

\newcommand\msmarco{MSMARCO NLG\xspace}

\newcommand\webnlg{WebNLG-en\xspace}
\newcommand\commongen{CommonGen\xspace}

\newcommand\rouges{R1 / R2 / RL\xspace}

\newcommand\minrisk{reward\xspace}
\newcommand\maxmargin{margin\xspace}
\newcommand\hinge{rank\xspace}
\newcommand\listrank{list rank\xspace}

\newcommand\ce{cross entropy\xspace}
\newcommand\kl{KL divergence\xspace}

\newcommand\bs{beam\xspace}
\newcommand\dbs{diverse beam\xspace}
\newcommand\ns{nucleus\xspace}

\newcommand\rouge{ROUGE\xspace}
\newcommand\ppl{perplexity\xspace}

\newcommand\bertscore{BERTScore\xspace}
\newcommand\seqtoseq{sequence-to-sequence\xspace}

\newcommand\fto{fine-tuned-only\xspace}
\newcommand\slc{SLiC\xspace}

\newcommand\seqx{\mathbf{x}}
\newcommand\seqy{\mathbf{y}}
\newcommand\seqyt{\bar{\mathbf{y}}}

\newcommand\seqyg{\hat{\mathbf{y}}}
\newcommand\embt{\bar{\mathbf{e}}}
\newcommand\embg{\hat{\mathbf{e}}}

\newcommand\lcal{L^{\mathrm{cal}}}
\newcommand\lreg{L^{\mathrm{reg}}}

\newcommand\bt{\mathbf{\theta}}

\newcommand\aryx{(\bar{y}_t | \bar{\mathbf{y}}_{t-1},\seqx)}
\newcommand\simdec{$s_\theta(\mathbf{y}, \hat{\mathbf{y}}, \mathbf{x})$\xspace}
\newcommand\simtok{$s_{tok}(\mathbf{y}, \hat{\mathbf{y}})$\xspace}

\usepackage{amsmath,amsfonts,bm}

\def\eqref#1{equation~\ref{#1}}

\def\1{\bm{1}}

\DeclareMathAlphabet{\mathsfit}{\encodingdefault}{\sfdefault}{m}{sl}
\SetMathAlphabet{\mathsfit}{bold}{\encodingdefault}{\sfdefault}{bx}{n}

\title{Calibrating Sequence likelihood Improves Conditional Language Generation}

\author{Yao Zhao \\ \texttt{\footnotesize yaozhaoyz@google.com} 
\And
Misha Khalman \\ \texttt{\footnotesize khalman@google.com} 
\And 
Rishabh Joshi \\ \texttt{\footnotesize rishabhjoshi@google.com} 
\And 
Shashi Narayan \\ \texttt{\footnotesize shashinarayan@google.com}
\And
Mohammad Saleh \\ \texttt{\footnotesize msaleh@google.com} 
\And
Peter J. Liu \\ \texttt{\footnotesize peterjliu@google.com} 
\AND \and 
\quad\quad\quad\quad\quad\quad\quad\quad\quad
Google Research, Brain Team
}

\begin{document}
\maketitle
\begin{abstract}

Conditional language models are predominantly trained with maximum likelihood estimation (MLE), giving probability mass to sparsely observed target sequences. 
While MLE trained models assign high probability to plausible sequences given the context, the model probabilities often do not accurately rank-order generated sequences by quality. 
This has been empirically observed in beam search decoding as output quality degrading with large beam sizes,
and decoding strategies benefiting from heuristics such as length normalization and repetition-blocking.
In this work, we introduce {\em sequence likelihood calibration} (\slc) where the likelihood of model generated sequences are calibrated to better align with reference sequences in the model's latent space.
With \slc, decoding heuristics become unnecessary and decoding candidates’ quality significantly improves regardless of the decoding method.
Furthermore, \slc shows no sign of diminishing returns with model scale, and presents alternative ways to improve quality with limited training and inference budgets.
With \slc, we exceed or match SOTA results on a wide range of generation tasks spanning abstractive summarization, question generation, abstractive question answering and data-to-text generation, even with modest-sized models.

\end{abstract}

\section{Introduction}

Conditional language generation aims to generate natural language text based on input context, and includes
 many useful and hard tasks such as abstractive summarization \citep{mani2001automatic,Nenkova:McKeown:2011},
generative question answering \citep{msmarco}, question generation \citep{nqg2017} and data-to-text \citep{wiseman-etal-2017-challenges,gardent-etal-2017-webnlg} tasks.
Pretraining large Transformer encoder-decoder models and fine-tuning them on downstream tasks is the common paradigm to address these tasks \citep{ t5, bart,ul2, pegasus}.

Conditional language generation tasks are modeled by learning the probability of a target sequence $\seqy$ given a context sequence $\seqx$.
Since directly modeling sequence probability $P(\seqy|\seqx)$ over all possible generated text sequences is intractable, the canonical solution is to auto-regressively factor the probability and share the parameters at all token prediction steps as 
$P_\bt(\seqy|\seqx) = \prod_{t=0}^l P_{\bt}(y^t | y^0 ... y^{t-1}, \seqx)$, where $l$ is the sequence length.
These models are often trained with maximum likelihood estimation (MLE) over observed target sequences. The learning objective thus becomes
$L = \sum_i^N -log(P_{\bt}(\seqy_i | \seqx_i)) =  \sum_i^N \sum_{t=0}^l -log(P_\bt(y_i^t | y_i^0 ... y_i^{t-1}, \seqx_i))$, where $N$ is the number of training instances.
It is also referred to as next token prediction loss as it is mathematically equivalent.

In the ideal setting of MLE training, a large number of target sequences are observed for each context, and the relative frequencies of output sequences can calibrate the assigned model probabilities.
However, in practice most language generation training datasets have only a single target sequence given the context.
While the subsequent MLE trained models learn to assign relatively high probability to plausible sequences, they lack the direct supervision to compare such sequences, and solely rely on models' generalization capability.
We refer to this phenomenon as models' sequence likelihood not being {\em calibrated}.
Prior works \citep{liu-liu-2021-simcls, liu-etal-2022-brio} has shown that the correlation between sequence probability and its quality for MLE trained models can be low. 
\citet{liu-etal-2022-brio} attributed this similarly as the deterministic (one-point) target distribution problem. 
Exposure bias \citep{ranzato_iclr16} further aggravates the problem, as sequence likelihood estimation is noisier when models' decoded sequences shift from exposed training data distribution. 

Many effective heuristics have been proposed during training and decoding to combat the problem of uncalibrated sequence likelihood.
Label smoothing \citep{DBLP:journals/corr/SzegedyVISW15} prevents the network from becoming over-confident towards the observed target.
This is particularly necessary in language generation, since the gold target represents just one of many possibilities.
It has been observed that increasing number of decoding candidates past a certain point leads to worse quality for beam search decoding \citep{yang-etal-2018-breaking, koehn-knowles-2017-six} and sampling \citep{adiwardana2020towards}.
An optimal number of decoding candidates is often determined empirically by decoding models on the validation set and measuring their performance.
Using length normalization is also essential for beam search decoding \citep{wu2016google} and sampling \citep{adiwardana2020towards} as models tend to underestimate sequence likelihood of longer sentences.
Repetition is another common failure mode when models overestimate the probability of repeated sequences \citep{holtzman2019curious}.
Tri-gram blocking \citep{paulus2018a} and nucleus sampling \citep{nucleus} have been used to interrupt repeating sequences.
These techniques are pervasive and often the default in modern Transformer libraries \citep{wolf-etal-2020-transformers, bart,t5,pegasus}.

Since the lack of observed target sequences in MLE training is the root problem, solutions involving learning with multiple sequence candidates have been proposed to directly address it.
They can be loosely put in three categories: (1) reinforcement learning with sequence-level rewards \citep{paulus2018a, ziegler2019fine, openai-human-feedback}; (2) two-stage systems that generate and rerank candidates \citep{liu-liu-2021-simcls, ravaut2022summareranker, liu-etal-2022-brio}; and (3) multi-task learning with sequence-level losses \citep{edunov-etal-2018-classical, liu-etal-2022-brio}.
Refer to Related Works (\autoref{sec:related_works}) for a more comprehensive discussion.

\begin{figure}[t!] %
 \begin{subfigure}[b]{0.55\textwidth}
    \centering
     \includegraphics[width=\textwidth]{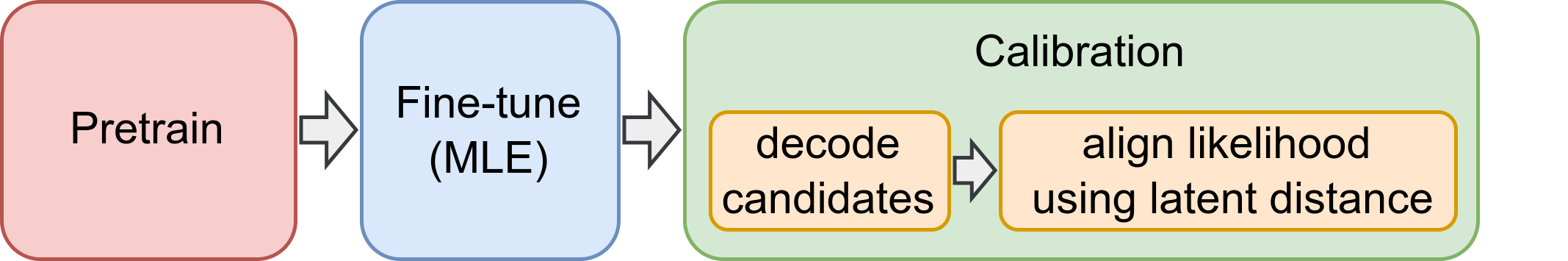}
     \vspace{12pt}

 \end{subfigure}
 \hfill
 \begin{subfigure}[b]{0.39\textwidth}
     \centering
     \includegraphics[width=\textwidth, trim=50 30 0 50]{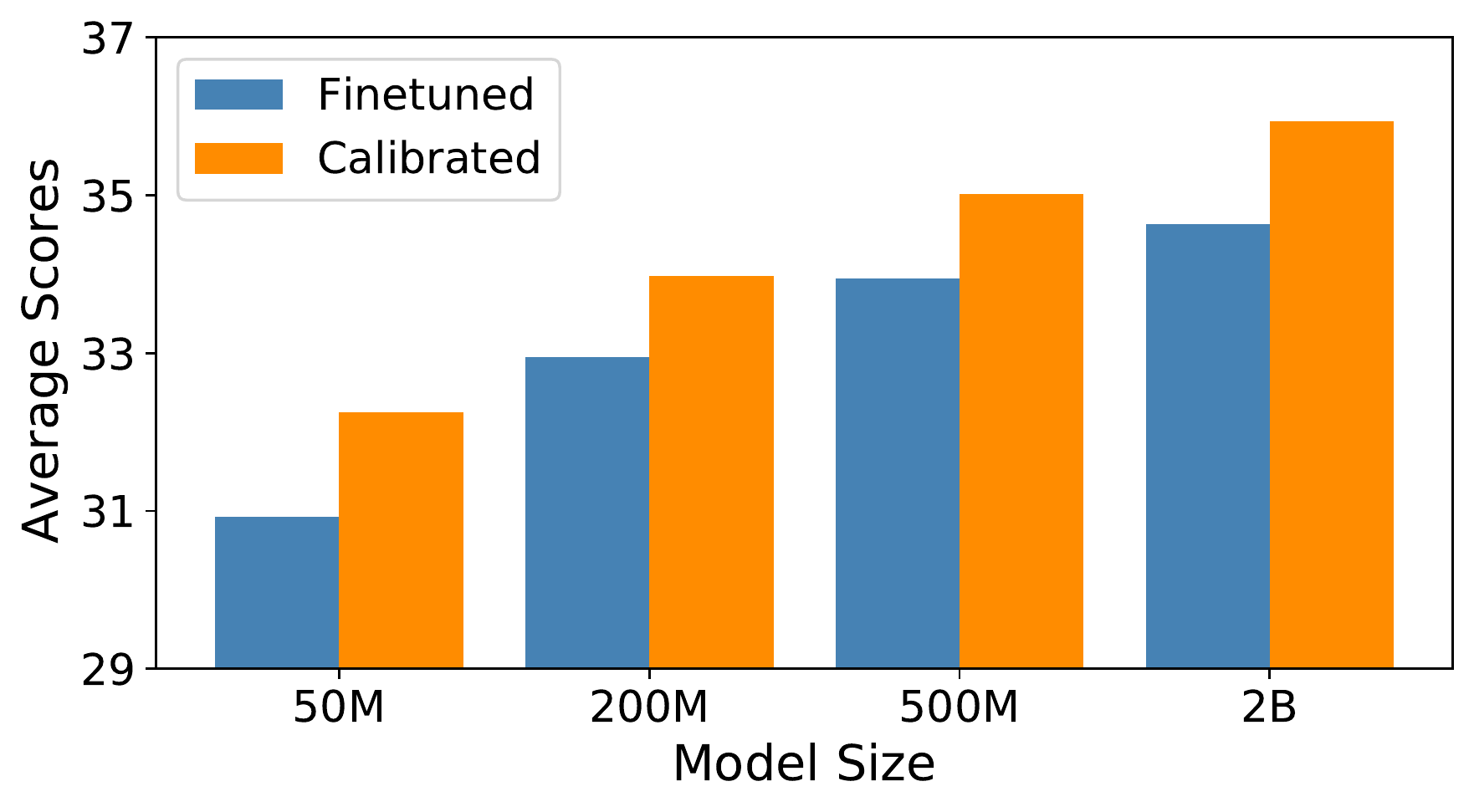}
 \end{subfigure}
\caption{Calibrating sequence likelihood improves language generation across model scales.
Scores are averaged \rouge across 4 datasets ($R_m$ in \autoref{sec:exp_details})
}
\label{fig:summary}
\end{figure}

In this paper, we propose to first decode candidates from a fine-tuned model on its own training dataset, and then continue training the model with a new objective.
The new objective aims to align candidates' sequence likelihoods according to their similarities to the target sequence in the model's latent space.  
We refer to this process as \textbf{sequence likelihood calibration} (\slc). Our approach is related to multi-task learning with sequence-level losses in \citet{liu-etal-2022-brio}.
However, we propose a simple yet effective recipe that eliminates decoding heuristics and doesn't risk directly optimizing the same metrics that are used to report text generation quality. Unlike reinforcement learning, it is a one-time offline process that avoids costly online decoding processes. Also, when compared to two-stage reranking systems, it doesn't require a separate reranking model that incurs additional complexity and compute. As depicted in \autoref{fig:summary}, our calibration stage naturally extends the current paradigm of pretraining and fine-tuning, and we show that calibrated models have strong improvements over \fto models across model sizes.

Our main contributions include:
\begin{itemize}

\item 
Proposed a sequence likelihood calibration (\slc) stage that consistently improves model quality, 
exceeding or matching state-of-the-art results on abstractive summarization, generative question answering, question generation and data-to-text generation tasks.

\item Proposed a novel calibration similarity metric between model decodes and targets measured in the model's latent space rather than resorting to external metrics or human feedback.

\item Demonstrated that \slc eliminates the need for popular decoding heuristics, such as beam size optimization, length normalization and repetition prevention for the calibrated models. 

\item Demonstrated that \slc has persistent significant benefits on model performance even as the number of model parameters scales up. Under the same inference budget, smaller calibrated models might outperform larger counterparts by decoding more candidates.

\end{itemize}

\section{Calibrating Sequence Likelihood}

We extend the common paradigm of pretraining and fine-tuning by introducing a third calibration stage, \slc.
As shown in \hyperref[alg:calibration]{Algorithm \ref*{alg:calibration}},
we first decode $m$ candidates $\{\seqyg\}_m$ from a fine-tuned model $P_{\bt_{ft}}(\seqy|\seqx)$ on fine-tuning dataset $\{\seqx, \seqyt\}_n$ and then calibrate the fine-tuned model by continuing training on our proposed loss:
$\mathcal{L}(\bt) = \sum_b \lcal(\bt, s; \seqx, \seqyt, \{ \seqyg \}_m) +\lambda \lreg (\bt, \bt_{ft}; \seqx, \seqyt)$
, where $\lcal$ and $\lreg$ are the calibration and regularization losses.
$s=s(\seqyg, \seqyt; \seqx)$ measures the similarity between the candidate $\seqyg$ and the target $\seqyt$ conditioned on the context $\seqx$.
We discuss choices of $s$, $\lcal$, $\lreg$ and decode strategies $ \seqyg \sim P_{\bt}(\seqy  | \seqx)$ in the following sections.

\begin{algorithm}
\caption{Calibrating Sequence Likelihood}
\label{alg:calibration}
\begin{algorithmic}
\For{ $\seqx, \seqyt \in \{ \seqx, \seqyt \}_n $ }
\Comment{sample $m$ candidates from the fine-tuned model}
\State  $\{ \seqyg \sim P_{\bt_{ft}}(\seqy | \seqx) \}_m $
\EndFor
\State $\bt \gets \bt_{ft}$
\Comment{initialized from the fine-tuned model}
\For{ $ \{ \seqx, \seqyt,  \{ \seqyg \}_m \}_b \sim  \{ \seqx, \seqyt,  \{ \seqyg \}_m \}_n $ }
\Comment{train with calibration and regularization loss}
\State $\bt \gets \bt - lr \nabla_{\bt}  \mathcal{L}(\bt)$ 
\EndFor
\end{algorithmic}
\end{algorithm}

\subsection{Similarity Function}
\label{sec:score_fn}
For a given output sequence $\seqy$, we take the decoder output hidden states $\mathbf{e}^{L\times D} = emb(\seqy, \seqx)$ as its representations, where $L$ is the number of tokens and $D$ is the hidden states dimension. 
Between a candidate $\seqyg$'s representations $\embg$ and the target $\seqyt$'s representations $\embt$,
we calculate their cosine similarities on spans of $n$ tokens and aggregate them across the sequences with a F-measured based function $F_n$. Notation of $F_n, P_n, R_n$ are same as in \bertscore \citep{bertscore}.

\begin{equation*} \label{eq:embedding_score}
\begin{split}
s_\bt(\seqyg, \seqyt; \seqx) = \sum_n F_n (\embg, \embt) 
= \sum_n F_n (emb(\seqyg, \seqx), emb(\seqyt, \seqx))
~~~~~~~~
F_n = 2\frac{P_n \times R_n}{P_n + R_n} \\
P_n(\embg, \embt) = \frac{1}{|\embg|} \sum_{\embg_{i:i+n}} \underset{\embt_{j:j+n}}{\max} ~ \embg_{i:i+n}^T \embt_{j:j+n}
 ~~~~~~~~
R_n(\embg, \embt) = \frac{1}{|\embt|} \sum_{\embt_{j:j+n}} \underset{\embg_{i:i+n}}{\max} ~ \embg_{i:i+n}^T \embt_{j:j+n} 
\end{split}
\end{equation*}

Compared to \bertscore, we use our models' decoder output representations instead of BERT encoder representations and also consider matching on spans of $n=1,2,4,8$ tokens rather than $1$.

Compared to using external metrics, such as \rouge, \bertscore,  this scoring function has a few advantages: 
(1) it adds very little compute cost, does not require extra model or out-of-graph computation; 
(2) it differs from the metrics that we evaluate the generation systems with and mitigates the risk of directly optimizing towards those imperfect metrics \citep{paulus2018a, openai-human-feedback}; 
(3) it is conditioned on the context $s(\seqyg, \seqyt; \seqx)$, as opposed to metrics in the form of $s(\seqyg, \seqyt)$.

\subsection{Calibration Loss}

The calibration loss $\lcal(\bt,s;\seqx,\seqyt,\{\seqyg\}_m)$ aims to align models' decoded candidates' sequence likelihood $P_\bt(\seqyg | \seqx)$ according to their similarity with the target sequence $s(\seqyg, \seqyt; \seqx)$.
Given the context $\seqx$, target $\seqyt$ and a set of candidates $\{\seqyg\}_m$, we consider the following 4 loss types.
\textbf{Rank} loss optimizes the ranking order of positive and negative candidates pairs $\seqyg_+, \seqyg_- $ uniformly sampled from $ \{\seqyg\}_m$ where $ s(\seqyg_+, \seqyt; \seqx) > s(\seqyg_-, \seqyt; \seqx)$.
\textbf{Margin} loss maximizes the sequence probability gap of positive and negative candidates pairs.
\textbf{List-wise rank} loss optimizes the ranking orders of a list of candidates, where $i,j$ are positions of $\seqyg_i,\seqyg_j$ in the set $\{\seqyg\}_m$ sorted by $s(\seqyg, \seqyt; \seqx)$. It is the contrastive loss used in BRIO \citep{liu-etal-2022-brio}.
\textbf{Expected reward} loss (or expected minimum risk) maximizes the expected similarity of a list of candidates \citep{edunov-etal-2018-classical}.

\begin{equation}
\label{eq:seq_loss}
\begin{split}
\lcal_{\mathrm{rank}} & =  \max(0, \beta - \log P_\bt(\seqyg_+ | \seqx) + \log P_\bt(\seqyg_- | \seqx)) \\
\lcal_{\mathrm{margin}} & =  \max(0, \beta(s(\seqyg_+, \seqyt; \seqx) -s(\seqyg_-, \seqyt; \seqx)) - \log P_\bt(\seqyg_+ | \seqx) + \log P_\bt(\seqyg_- | \seqx)) \\
\lcal_{\mathrm{list~rank}} & = \Sigma_{i < j} \max \left(0, \beta |i-j| - \log P_\bt(\seqyg_i | \seqx) + \log P_\bt(\seqyg_j | \seqx)\right)
\\
\lcal_{\mathrm{reward}} & = \Sigma_i \left[-s(\seqyg_i, \seqyt; \seqx) * \frac{P_\bt(\seqyg_i | \seqx)}{\sum_i P_\bt(\seqyg_i | \seqx)} \right]
\end{split}
\end{equation}
$\beta$ values for all losses are chosen empirically for each loss type in \autoref{sec:ablation}.

\subsection{Regularization Loss}
We consider two alternate types of regularization loss $\lreg$  to prevent models from deviating significantly from their fine-tuned MLE objective:
\textbf{Cross entropy} is the standard fine-tuning MLE objective used in \citep{liu-etal-2022-brio}.
\textbf{KL divergence} directly minimizes the probability distribution distance between the calibrated model and the fine-tuned model at each token on observed target sequence.
The regularization losses are both on token level.

\begin{equation}
\label{eq:reg_loss}
\lreg_{\mathrm{ce}} =  \sum_t - \log P_\bt\aryx
~~~~~
\lreg_{\mathrm{kl}} = \sum_t P_\bt\aryx \log \frac{P_\bt\aryx}{P_{\bt_{ft}}\aryx} \\
\end{equation}

\subsection{Candidates Decoding Methods}

We consider the following decoding methods for \slc:

\textbf{Beam Search} is the standard best-first algorithm to solve the intractable maximum likelihood optimization for \seqtoseq models  \citep{tillmann-ney-2003-word,li-etal-2016-deep,wiseman-etal-2017-challenges,chen-etal-2018-recurrent}.

\textbf{Diverse Beam Search} (DBS; \citeauthor{dbs}, \citeyear{dbs}) generates  a list of diverse outputs by dividing the beam search budget into groups and enforcing dissimilarity between groups of beams. It strikes balance between quality and diversity and is often the best strategy for two-stage reranking systems \citep{liu-liu-2021-simcls, ravaut2022summareranker, liu-etal-2022-brio}.

\textbf{Nucleus Sampling} \citep{nucleus} only samples high-probable tokens within cumulative probability $p$ at each step of the decoding. It produces diverse candidates while preventing sampling very low quality ones.

\section{Experiments}
\label{sec:experiments}

\subsection{Tasks and Datasets}

For abstractive summarization tasks, we choose 
\textbf{\cnn} \citep{hermann2015cnndm, see-etal-2017-get}, 
\textbf{\xsum} \citep{narayan-etal-2018-dont},
\textbf{\reddit} \citep{kim-etal-2019-reddittifu} and 
\textbf{\samsum} \citep{gliwa-etal-2019-samsum} due to their %
diversity in domain, style, abstractiveness, and summary lengths.
For question answering related tasks, we choose generative question answering given context
\textbf{\msmarco} \citep{msmarco} and its reverse problem of question generation
\textbf{\qg} \citep{nqg2017, du-etal-2017-learning} .
For data-to-text tasks, we choose text generation given structured data
\textbf{\webnlg} \citep{gardent-etal-2017-webnlg} and common concepts reasoning 
\textbf{\commongen} \citep{lin-etal-2020-commongen}.
More details of datasets can be found at \autoref{appendix:dataset} along with their statistics.

\subsection{Model training and evaluation details}
\label{sec:exp_details}
We follow the PEGASUS pretraining \citep{pegasus} and extend transformer model sizes to
\pegasussmall (\smallmodelsize), \pegasusbase (\basemodelsize), \pegasuslarge (\largemodelsize) and \pegasusxxlarge (\xxlargemodelsize).\footnote{Approximated size, accurate sizes are reported in \autoref{appendix:model_size}.} Different from the original paper, we use a sentencepiece 96k vocabulary with byte-fallback \citep{kudo2018sentencepiece} and pretraining batch size of $4096$ across all models. See \autoref{appendix:model_size} for model dimensions.

In all experiments, we use learning rate $lr=10^{-4}$, and batch sizes of 512 to finetune and 64 to calibrate models.
We use beam search to generate calibration candidates and evaluate the calibrated models, unless specified otherwise.

In our ablation studies (\autoref{sec:ablation}), benefits analysis (\autoref{sec:benefit}), and scaling experiments (\autoref{sec:scaling}), we use models pretrained to \num{500000} steps and conduct experiments on 4 datasets (\cnn, \xsum, \reddit and \samsum).
For ablation studies and benefits analysis, we use \pegasuslarge.
We report \rouge 1/2/L \citep{lin-2004-rouge}\footnote{Using pypi package \texttt{rouge-score}.
We report \texttt{rougeLsum} for ROUGE-L.} 
for each dataset on \textbf{validation} splits and their overall score $R_m$ defined as geometric mean of \rouge 1/2/L averaged across datasets, $R_m=\frac{1}{4} \sum_d \sqrt[3]{R_1R_2R_L}$.

For the final results (\autoref{sec:final}), we pretrain \pegasusxxlarge model to $2.5M$ steps, fine-tune it on all 8 datasets, calibrate them using the same recipe and report numbers on the \textbf{test} split (unless specified otherwise). We use corresponding standard evaluation scripts for each dataset.\footnote{For summarization datasets, we use pypi package \texttt{rouge-score}. For \qg and \msmarco, we use the original evaluation scripts provided by \citet{du-etal-2017-learning} and \citet{msmarco}, respectively. For \webnlg and \commongen, we use the versions from the GEM 
benchmark \citep{gehrmann-etal-2021-gem} and report using the GEM evaluation framework.
Those scripts mainly differ in text tokenization methods.}

\subsection{Ablation Studies of Calibration}
\label{sec:ablation}
Ablation experiment results discussed below can be found in \autoref{tab:loss_ablation}.

\begin{table}[tbh]
\caption{
Ablation of the sequence likelihood calibration method.
Shared hyper-parameters are held as constant within each comparison group but vary between groups (\autoref{appendix:ablation}).
$\Delta$ is the relative improvements of overall score $R_m$ compared with the fine-tuned model.
}
\begin{center}

\small
\setlength{\tabcolsep}{4pt}
\begin{tabular}{ll cccc c}
\hline
\multicolumn{2}{l}{Ablation}  & {\cnn} & {\xsum} & {\reddit} & {\samsum} & $\Delta$ \\
 &  & \rouges & \rouges & \rouges & \rouges  & avg \\ 
\hline
 & fine-tuned & 
44.74/21.83/41.92 & 47.23/24.31/39.12 & 26.84/9.08/21.92 & 53.67/29.35/44.75 & 0.00\%
\\
\hline

\multicolumn{7}{l}{\emph{similarity function }} \\

& ROUGE & 
46.47/22.49/43.63 & 47.86/24.55/39.58 & 29.92/9.83/23.93 & 54.82/30.15/45.30 & 3.26\%
\\
& decoder repr & 
46.55/22.50/43.69 & 47.88/24.62/39.62 & 29.86/9.84/23.91 & 54.72/29.96/45.10 & 3.20\%
\\
& token emb & 
46.51/22.48/43.67 & 47.04/23.63/38.39 & 29.78/9.69/23.46 & 53.71/29.38/44.75 & 1.64\%
\\
\hline

\multicolumn{7}{l}{\emph{calibration loss}} \\
& \hinge & 
46.73/22.70/43.85 & 48.11/24.80/40.06 & 30.34/9.80/24.32 & 55.19/30.46/46.32 & 4.27\%
\\
& \maxmargin &
46.11/22.46/43.30 & 47.62/24.81/39.89 & 30.84/9.97/24.37 & 54.58/30.10/45.92 & 3.63\%
\\
& \listrank &
46.62/22.88/43.76 & 47.93/24.57/39.67 & 30.87/9.65/24.46 & 54.56/29.81/45.17 & 3.49\%
\\
& \minrisk & 
46.49/22.55/43.63 & 47.77/24.48/39.49 & 30.99/9.95/24.39 & 54.42/29.98/45.56 & 3.47\%
\\
\hline

\multicolumn{7}{l}{\emph{regularization loss} } \\
& none &
46.54/22.44/43.68 & 47.51/24.70/39.82 & 30.73/9.68/24.05 & 55.07/30.07/45.60 & 3.48\%
\\
& cross entropy &
46.73/22.70/43.85 & 48.11/24.80/40.06 & 29.96/9.72/23.82 & 55.19/30.46/46.32 & 4.06\%
\\
& KL divergence & 
46.80/22.83/43.98 & 47.96/24.92/40.09 & 30.73/9.68/24.05 & 54.87/30.20/45.95 & 4.09\%
\\
\hline

\multicolumn{7}{l}{\emph{candidates decoding method}} \\
& beam search & 
46.50/22.48/43.66 & 47.82/24.65/39.67 & 31.04/9.96/24.37 & 54.66/30.27/45.46 & 3.70\%
\\
& diverse beam &
46.31/22.48/43.47 & 47.79/24.53/39.51 & 31.00/9.95/24.08 & 54.57/29.67/45.55 & 3.26\%
\\
& nucleus & 
46.45/22.46/43.54 & 47.67/24.50/39.47 & 31.09/10.01/24.31 & 54.61/30.04/45.63 & 3.51\%
\\
\hline

\multicolumn{7}{l}{\emph{calibration checkpoint selection}} \\
& ROUGE & 
46.66/22.66/43.84	& 48.03/24.78/39.79 & 30.94/9.98/24.43 & 54.63/30.03/45.79 & 3.96\%
\\
& perplexity & 
47.36/24.02/44.45 & 47.96/24.74/39.78 & 31.04/10.08/24.53 & 54.65/30.11/46.00 & 4.93\%
\\
\hline

\end{tabular}
\end{center}

\label{tab:loss_ablation}
\end{table}

\textbf{Similarity Function}
We compare our proposed similarity function, using models' latent states at decoder output representation $s_\bt(\seqyg, \seqyt; \seqx)$ (\autoref{sec:score_fn}),
to directly optimizing the evaluation metric \rouge.
They perform similarly on all datasets even when evaluation metrics are \rouge scores.
We also test a variant of our similarity function by replacing decoder representation $emb(\seqy, \seqx)$ with token embeddings.
This variant has lower performance, which suggests benefits of contextualized and input-dependent representations.

\textbf{Calibration Loss}
Calibrated models with all loss types significantly improve over \fto models. Rank loss performs the best followed by \maxmargin, \listrank and then \minrisk. Reward maximization has the advantage of no hyper-parameters $\beta$ (\autoref{eq:seq_loss}) to sweep while \hinge and \maxmargin loss have smaller training memory footprints. 
Rank loss showing the best gain indicates that relative ordering of candidates is more important than the absolute value of their similarity to the target.

\textbf{Regularization Loss}
Cross entropy and KL divergence regularization perform similarly. About 85\% of the calibration gain remains if regularization is removed.

\textbf{Calibration Candidates Decoding Method}
We choose hyper-parameters for calibration candidates decoding methods based on validation set.
The optimal decoding method is dataset dependent, however the differences between methods are small and the worst method achieves 90\% of the gains of the best one.
Beam search yields the highest average quality.
This is opposite to the findings in the two-stage reranking systems \citep{liu-liu-2021-simcls, ravaut2022summareranker, liu-etal-2022-brio}, where more diverse decoding strategies are preferred.

\textbf{Checkpoint Selection for Fine-tuned Model}
We compare ROUGE-selected and perplexity-selected checkpoints.
The experiments show that starting calibration from the perplexity-selected checkpoint yields same or better performance with the biggest gap on \cnn dataset.

\tldr We recommend a simple recipe: select the fine-tuned model's checkpoint by its validation set perplexity; decode candidates using beam search; calibrate the model with rank loss and KL divergence regularization.

\subsection{Benefits of Calibrated Sequence Likelihood}
\label{sec:benefit}

\begin{figure}[bth]
 \centering
 \begin{subfigure}[t]{0.21\textwidth}
     \centering
     \caption*{\samsum}
     \includegraphics[width=\textwidth, trim=30 0 0 30]{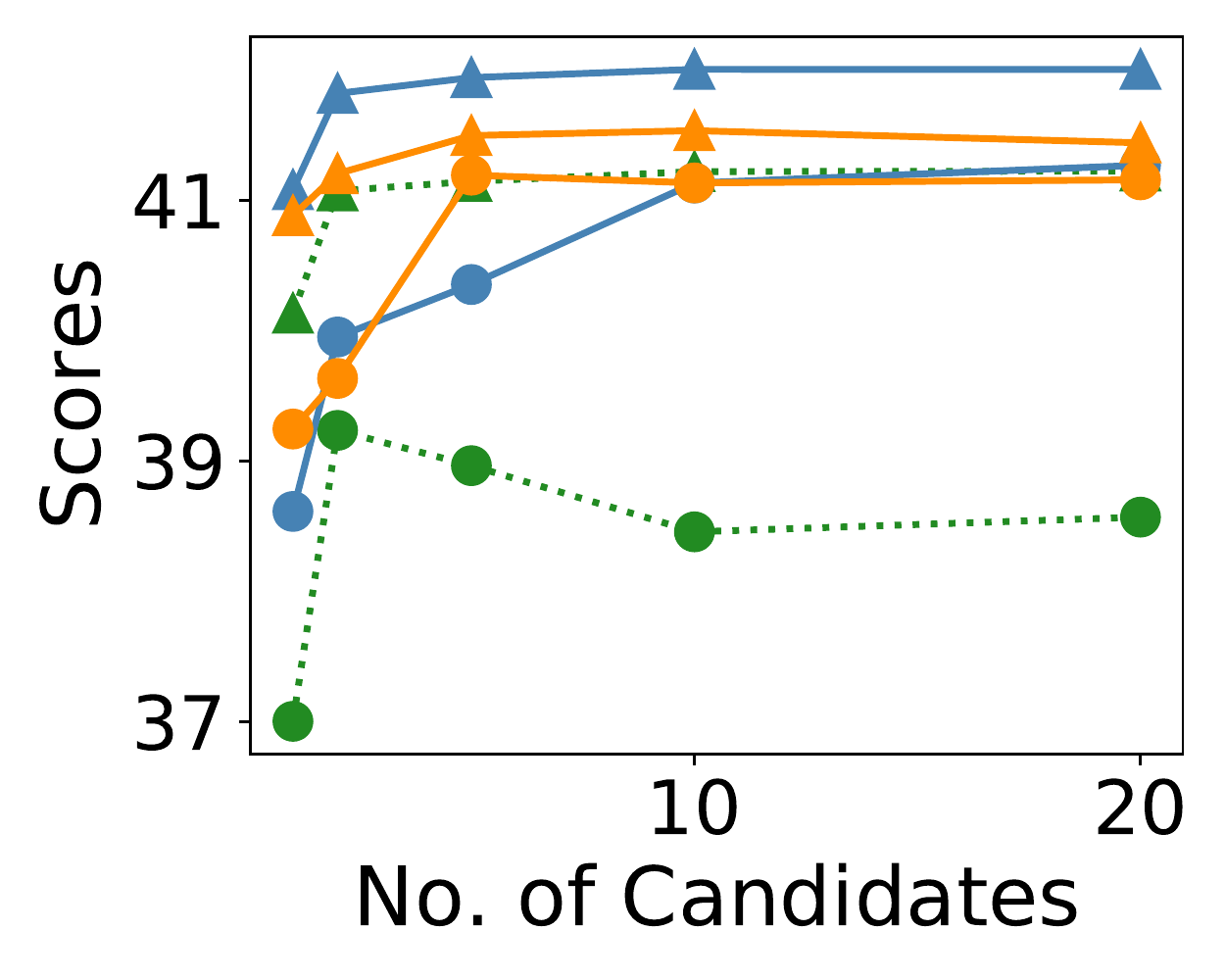}
 \end{subfigure}
 \hfill
 \begin{subfigure}[t]{0.21\textwidth}
     \centering
     \caption*{\reddit}
     \includegraphics[width=\textwidth, trim=30 0 0 30]{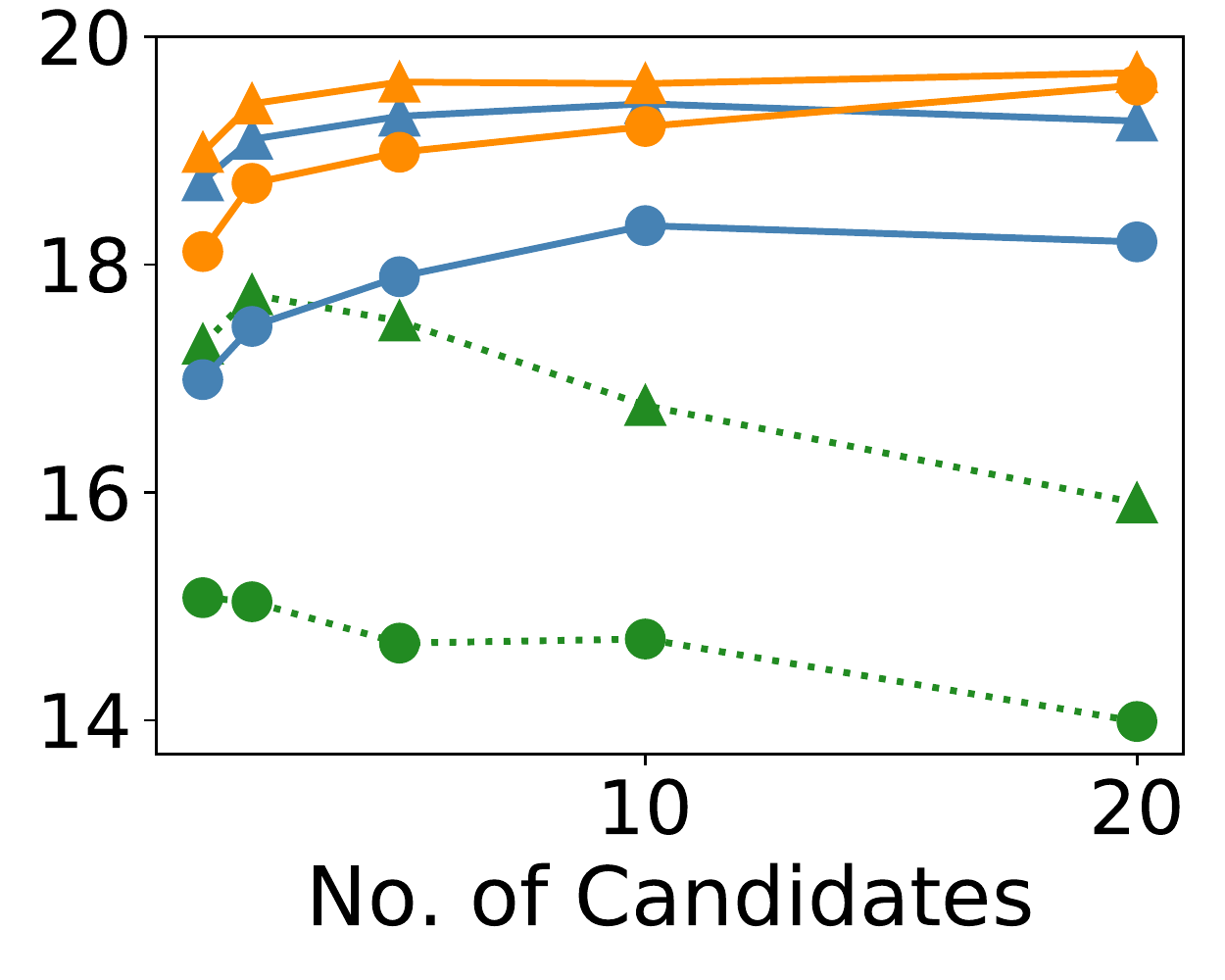}
 \end{subfigure}
 \hfill
 \begin{subfigure}[t]{0.21\textwidth}
     \centering
     \caption*{\cnn}
     \includegraphics[width=\textwidth, trim=30 0 0 30]{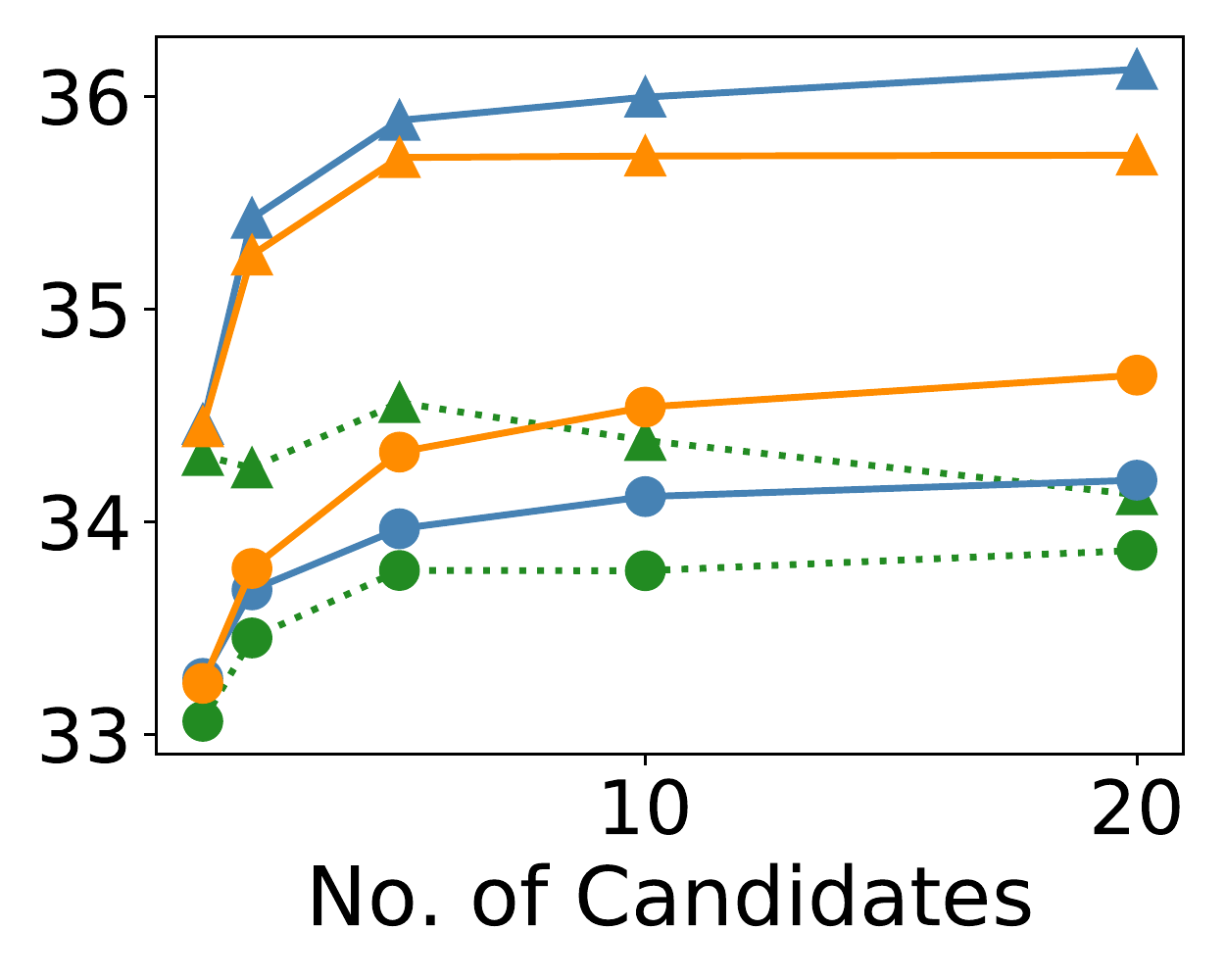}
 \end{subfigure}
 \hfill
 \begin{subfigure}[t]{0.21\textwidth}
     \centering
     \caption*{\xsum}
     \includegraphics[width=\textwidth, trim=30 0 0 30]{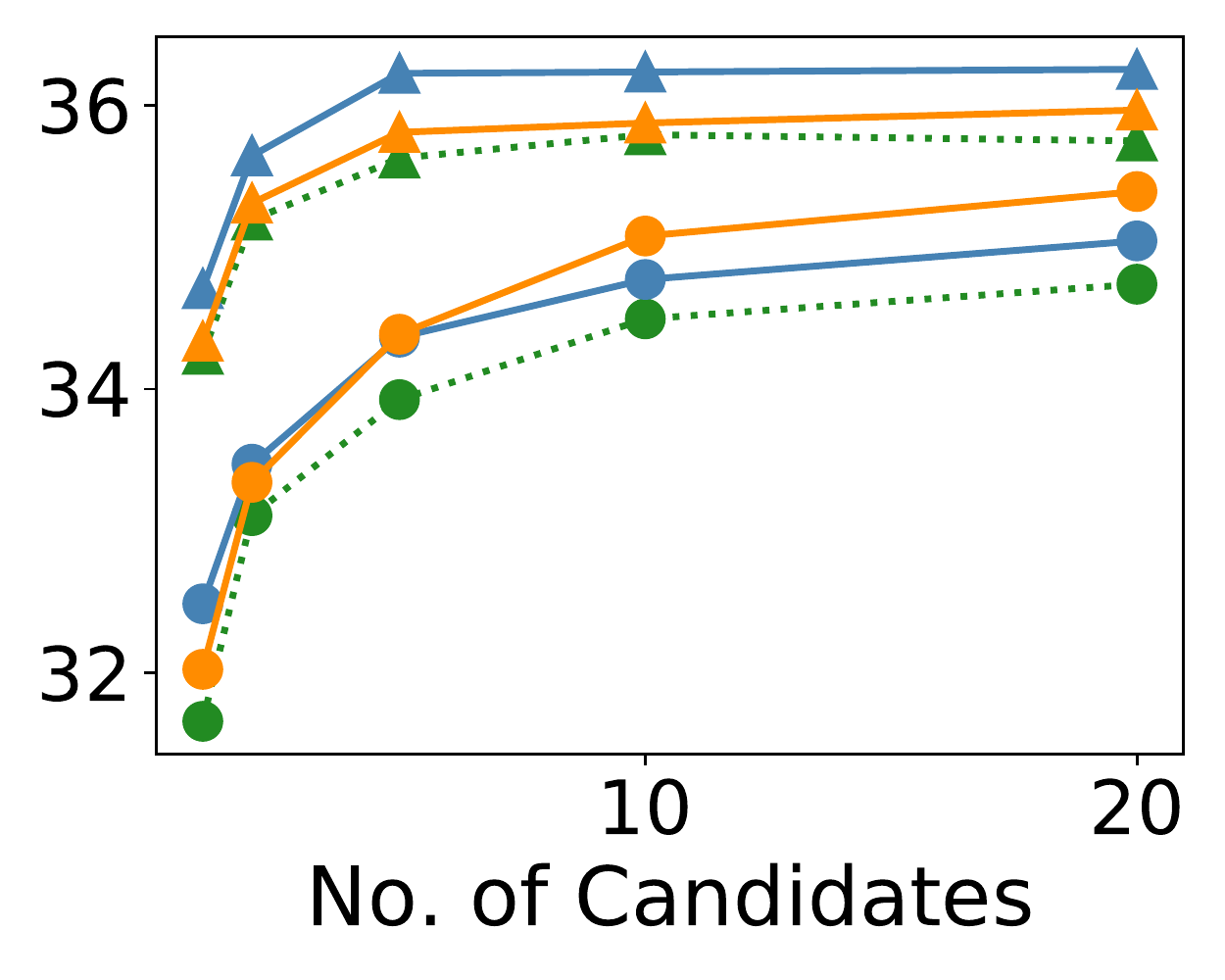}
 \end{subfigure}
 \hfill
 \begin{subfigure}[t]{0.12\textwidth}
     \centering
     \caption*{}
     \includegraphics[width=\textwidth, trim=50 0 50 30]{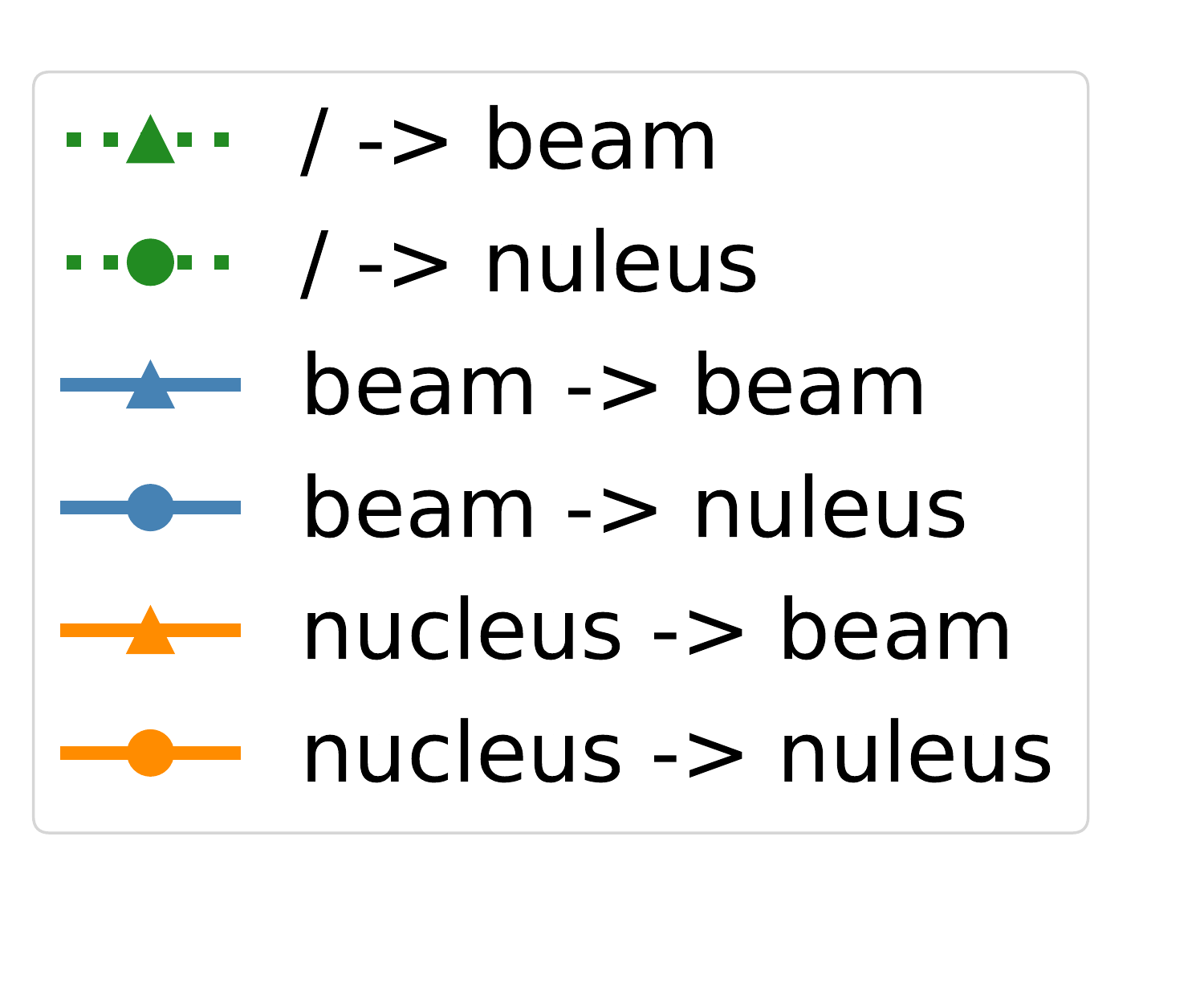}
 \end{subfigure}
 \hfill

\caption{
Effect of decoding methods on calibrated and fine-tuned only models.
Colors indicate calibration method.
Markers indicate evaluation decoding method.
Hyper-parameters at \autoref{appendix:decoding_method}.}
\label{fig:decoding}
\end{figure}

\textbf{Calibrated models' quality monotonically improves as the number of decoding candidates increase,\footnote{ At evaluation-decoding time, the candidate with the highest sequence probability is selected to compute quality for both beam search and nucleus sampling.}} regardless of the calibration-decoding and evaluation-decoding methods, as shown in \autoref{fig:decoding}.
On the other hand, \fto models suffer from decreased quality when the number of decodes exceeds an optimal value.
Once a model is calibrated with either decoding method, it performs well with both at evaluation time. 
Decoding with beam search yields higher scores, verified up to 20 decodes.
When the calibration-decoding and the evaluation-decoding method align, the final quality is slightly better than the mismatched settings.
\cnn, \xsum, and \samsum datasets work best with beam search, however \reddit works better with nucleus sampling and decoding it with a larger number of candidates may achieve better results.

\textbf{Calibrated models do not require length normalization.}
As shown in \autoref{tab:brevity_penalty}, length normalization (commonly implemented as $\alpha$ for beam search) is essential for \fto models which bias towards longer sequences at decoding time. In contrast, length normalization has minimal effect on calibrated models.

\textbf{Calibrated models suffer from far fewer repetitions.}
The repetition rate (rep\%) measures a common mode of model failures.
It is defined as the percentage of examples that contain any kind of consecutive repeated word n-grams, 
While length normalization helps general quality on the \fto models, it leads to  a side-effect of  higher repetitions.
Calibrated models, with or without length normalization, have a much lower repetition rate.
When we compare with the repetition rate in the gold reference (repetition may occur naturally), calibrated models without length normalization have similar or lower repetition rate.
\begin{table}[tbh]
\caption{
Comparison between fine-tuned only models and calibrated models with or w/o brevity penalty $\alpha$ on overall quality (\rouges) and repetitions' occurrence percentage (rep\%). 
Hyper-parameters at \autoref{appendix:brevity_penalty}. 
}
\centering
\begin{center}

\setlength{\tabcolsep}{4pt}

\scriptsize
\begin{tabular}{ll ccccccccc}
\hline
\slc & $\alpha$ & \multicolumn{2}{c}{\cnn} & \multicolumn{2}{c}{\xsum} & \multicolumn{2}{c}{\reddit} & \multicolumn{2}{c}{\samsum} & $\Delta$ \\
 & & \rouges & rep\% & \rouges &rep\% & \rouges & rep\% & \rouges & rep\% & avg \\ 
\hline
\multicolumn{2}{l}{gold reference} & - & 0.03 & - & 0.01 & - & 0.09 & & 0.05
\\
\xmark & \xmark & 
39.37/19.67/36.89 & 0.03 & 46.96/24.29/39.19 & 0.03 & 26.62/8.91/21.77 & 0.26 & 50.28/27.25/42.69 & 0.00 & -5.15\%
\\
\xmark & \cmark &
44.74/21.83/41.92 & 0.13 & 47.23/24.31/39.12 & 0.07 & 26.84/9.08/21.92 & 0.90 & 53.67/29.35/44.75 & 0.20 & 0.00\%
\\
\cmark & \xmark &
46.44/22.38/43.57 & 0.02 & 47.57/24.42/39.46 & 0.03 & 30.99/9.95/24.39 & 0.03 & 54.42/29.98/45.56 & 0.00 & 3.31\%
\\
\cmark & \cmark &
46.49/22.55/43.63 & 0.03 & 47.77/24.48/39.49 & 0.03 & 30.98/9.96/24.30 & 0.12 & 54.64/30.01/45.17 & 0.08 & 3.42\%
\\
\hline

\end{tabular}

\end{center}

\label{tab:brevity_penalty}
\vspace{-10pt}
\end{table}

\tldr Calibrated models do not require decoding heuristics such as beam size optimization, length normalization and repetition blocking.

\subsection{Scaling Properties of Calibrated Models}
\label{sec:scaling}

\begin{figure}[bth]
 \centering

 \begin{subfigure}[t]{0.35\textwidth}
     \centering
     \caption*{\samsum}
     \includegraphics[trim={15 0 0 20}, width=\textwidth]{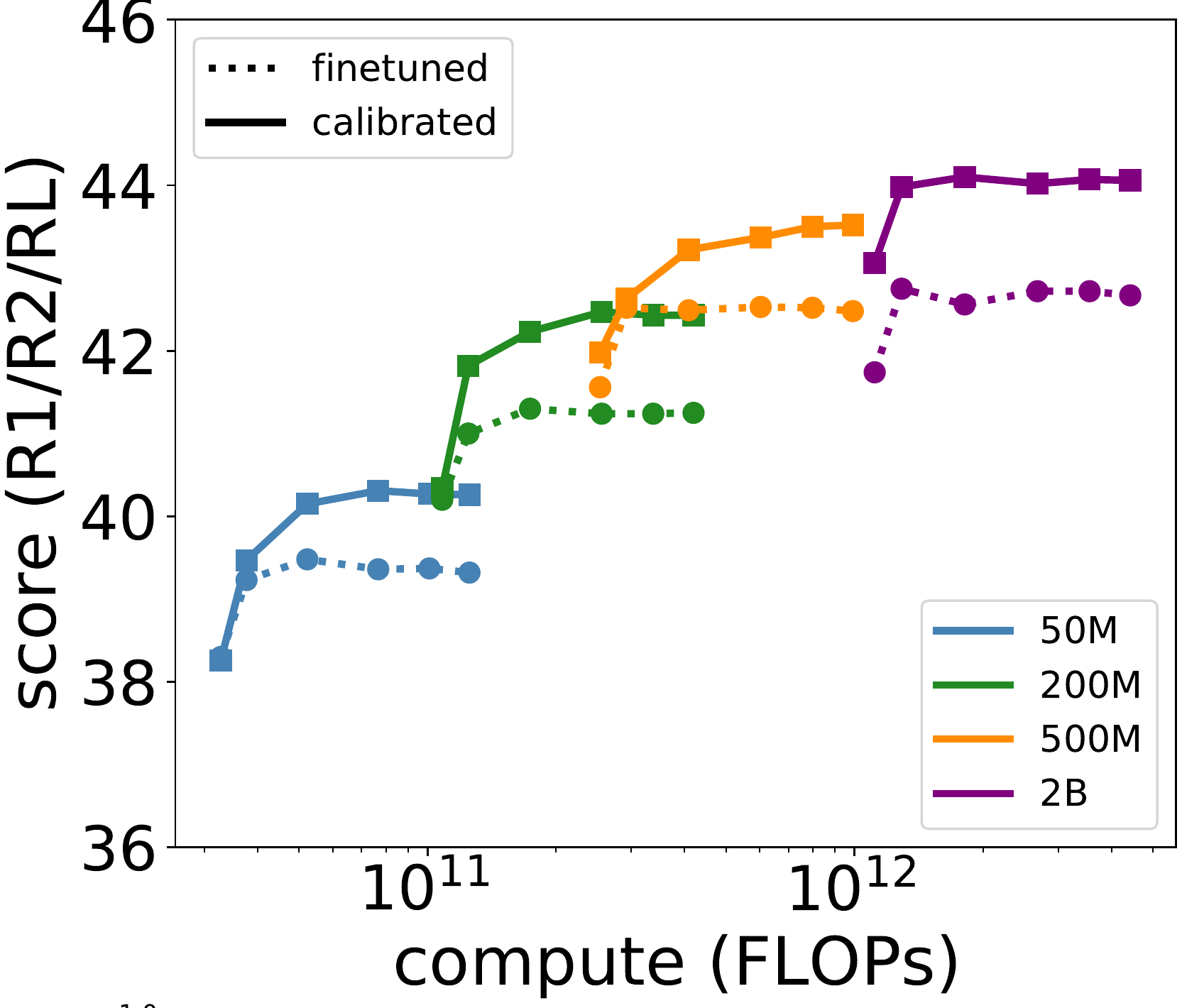}
 \end{subfigure}
 \hspace{15pt}
 \begin{subfigure}[t]{0.35\textwidth}
     \centering
     \caption*{\reddit}
     \includegraphics[trim={0 0 0 20}, width=\textwidth]{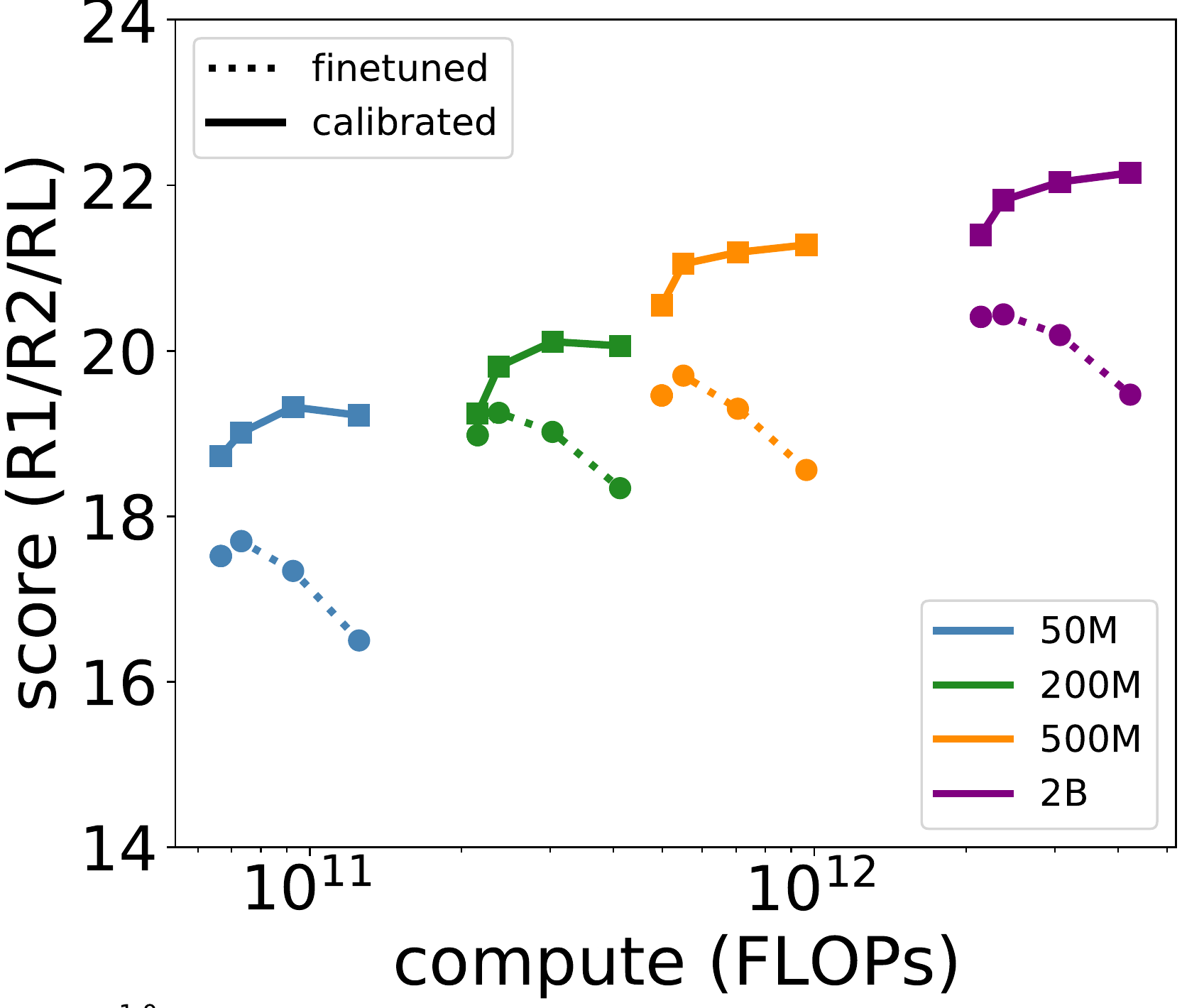}
 \end{subfigure}
 \hfill
 \\
 \vspace{15pt}
 \begin{subfigure}[t]{0.35\textwidth}
     \centering
     \caption*{\cnn}
     \includegraphics[trim={15 0 0 20}, width=\textwidth]{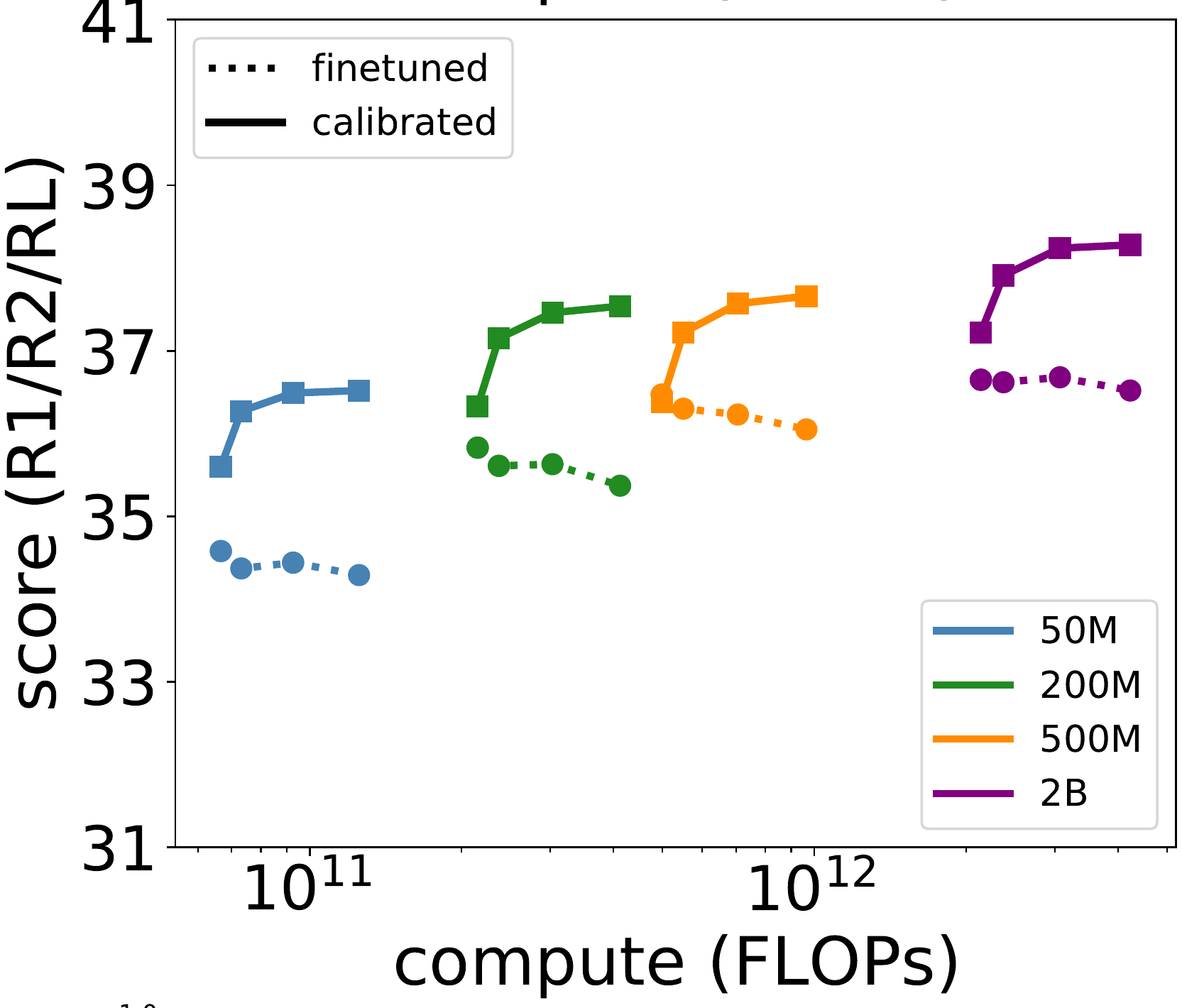}
 \end{subfigure}
 \hspace{15pt}
 \begin{subfigure}[t]{0.35\textwidth}
     \centering
     \caption*{\xsum}
     \includegraphics[trim={0 0 0 20}, width=\textwidth]{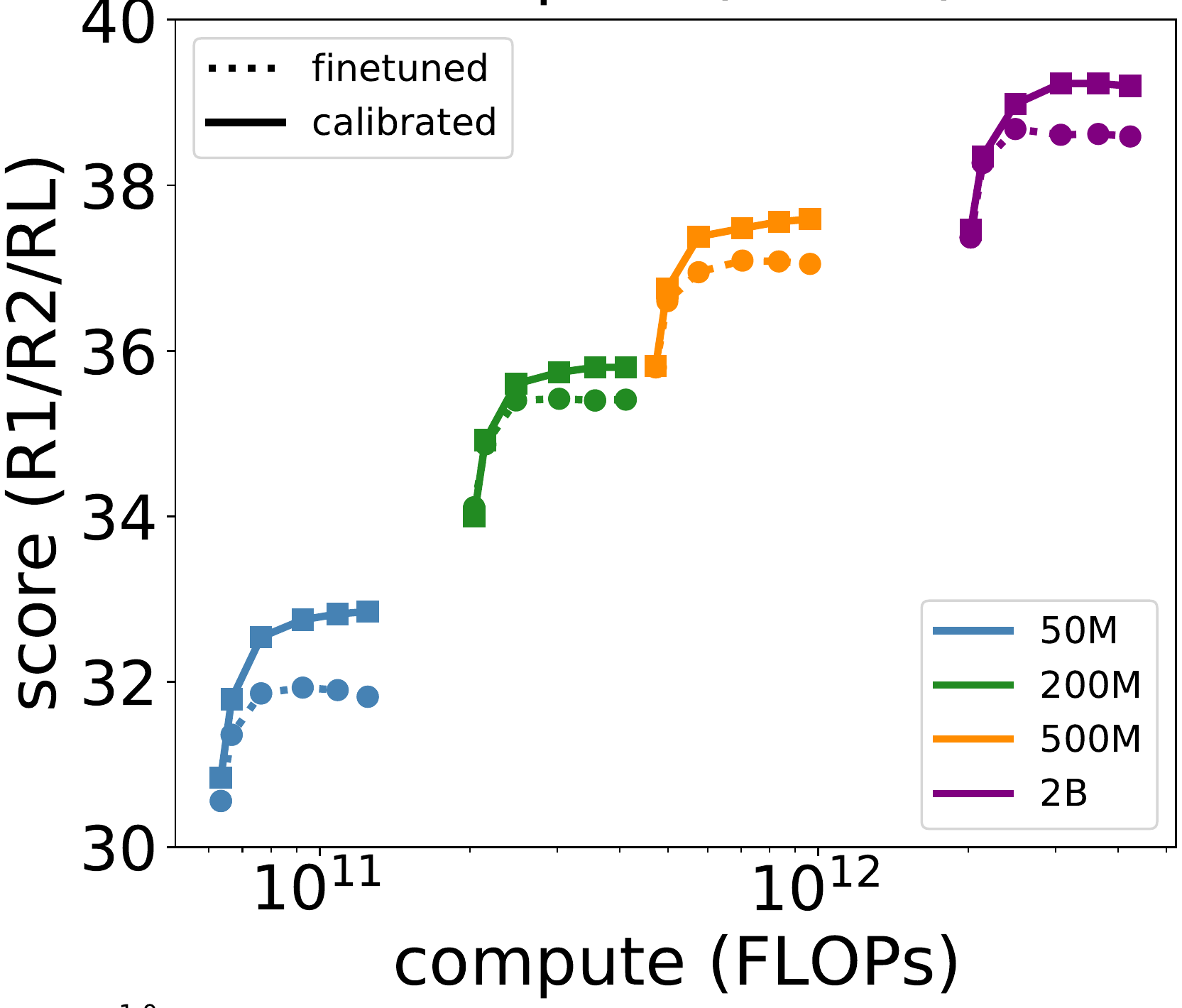}
 \end{subfigure}
 \hfill
 
\caption{
Quality and inference compute trade-off comparison between fine-tuned only and calibrated models.
Inference compute is scaled by increasing model parameters (different colors) and number of decoding candidates (dots on the same line).
Hyper-parameters at \autoref{appendix:scaling}.
}
\label{fig:scaling}
\end{figure}

Scaling properties are important for projecting a technique's future relevance as models scale up \citep{kaplan2020scaling}.
In \autoref{fig:scaling}, we compare generation quality versus inference compute at different model sizes and number of decoding candidates using beam search.
\autoref{appendix:flops} describes the method to estimate inference compute FLOPs.

As mentioned earlier in \autoref{sec:benefit}, \fto models have optimal decoding beam sizes while calibrated models' performance monotonically increase with larger decoding beam sizes.
Even in the case of greedy decoding (beam size of 1), the calibrated models' performance exceeds the \fto models, by a large margin for some datasets (\cnn and \reddit).
Their gaps grow larger with increasing number of beam sizes.

\textbf{The magnitude of quality improvement from calibration persists over models sizes spanning from \smallmodelsize to \xxlargemodelsize}.
There is no obvious sign of diminishing return as model size scales up.

\textbf{Inference compute may be used for decoding rather than on larger models.}
A calibrated model, once trained, can improve its performance by decoding more candidates, usually more effectively in the beginning, although returns diminish over 10 candidates.
In some cases (\samsum and especially \cnn), a smaller model decoding more candidates can beat a larger one at both quality and efficiency.

\tldr Calibration benefits persist as model sizes scale up.
Smaller calibrated models can outperform larger ones under the same inference compute budget.

\subsection{Final Results}
\label{sec:final}
We calibrate the fine-tuned \pegasusxxlarge models on 8 language generation tasks using the simple recipe identified in \autoref{sec:ablation} and evaluate them with beam search without decoding heuristics (\autoref{sec:benefit}).
The only hyper-parameter we optimize for \slc is learning rate $lr$ (\autoref{appendix:final}).
We use beam size 5 for \fto models and 10 for calibrated models.

As shown in \autoref{tab:final}, calibrated models show consistent improvement over \fto models across datasets and tasks. 
Overall, our calibrated models exceed or match the SOTA models on all datasets.
On \xsum, \samsum, \webnlg and \commongen, our calibrated \xxlargemodelsize models are ten to a hundred times smaller than the SOTA models.

\begin{table}[tbh]
\caption{Calibrated \pegasusxxlarge comparing with prior SOTA results:
BRIO\textsuperscript{a}\citep{liu-etal-2022-brio},
ULL\textsuperscript{b}\citep{ul2},
ST-MoE\textsuperscript{c}\citep{zoph2022designing},
UniLMv2\textsuperscript{d}\citep{bao2020unilmv}, 
Masque\textsuperscript{e}\citep{nishida-etal-2019-multi}, and BART+R3F\textsuperscript{f}\citep{aghajanyan2021better}.
$\dagger$ is on validation set.
* is on unknown split.
See hyper-parameters in \autoref{appendix:final}.
}
\begin{center}
\setlength{\tabcolsep}{6pt}
\small
\begin{tabular}{c cc cc}
\hline
Dataset & \multicolumn{2}{c }{Prior SOTA } &  Our fine-tuned (2B) & Our calibrated (2B) \\
& \#params & \rouges & \rouges & \rouges \\
\hline
\cnn 
& \bartmodelsize\textsuperscript{a} & 47.78/23.55/44.57
& 44.31/21.91/41.41 &
47.97/24.18/44.88 %
\\
\xsum 
& \stmoemodelsize\textsuperscript{c} & -- /27.1/--
& 49.57/26.77/41.41 & 
49.77/27.09/42.08	%
\\
\reddit &  
\bartmodelsize\textsuperscript{f} & 30.31/10.98/24.74\textsuperscript{*}
& 28.73/10.12/23.24 & 
32.03/11.13/25.51 %
\\
\samsum &
\ullmodelsize\textsuperscript{b} & --/29.60/--
& 53.64/29.21/44.83 &
54.37/29.88/45.89 %
\\
\hline
\qg &
\unilmmodelsize\textsuperscript{d} 
&  --/--/52.13 
& --/--/52.59 
& --/--/53.28
\\
\msmarco\textsuperscript{$\dagger$}
& UNK\textsuperscript{e} 
& --/--/69.77 
&  --/--/70.73 
& --/--/71.06
\\
\hline
\webnlg &
\ullmodelsize\textsuperscript{b} & --/55.40/-- &
76.96/52.97/62.56 
& 78.09/55.52/65.06

\\
\commongen\textsuperscript{$\dagger$} &
\ullmodelsize\textsuperscript{b} & --/37.40/-- 
& 66.49/36.17/58.82 
& 68.95/38.49/60.13

\\
\hline

\end{tabular}
\end{center}
\label{tab:final}
\end{table}

\tldr \pegasusxxlarge achieves SOTA results on a wide range of language generation tasks using a simple \slc recipe while eliminating decoding heuristics.

\section{Related Works}
\label{sec:related_works}

\subsection{RL approaches}

\citet{paulus2018a} directly optimizes evaluation metric ROUGE in RL fine-tuning stage.
One issue is that ROUGE metric does not enforce fluency. 
The authors found summaries to be not always readable and proposed that using a mixed training objective works better.

\citet{ziegler2019fine, openai-human-feedback} collects human judgements on fine-tuned models' decodes to train a reward model that ranks candidates according to human preferences.
The supervised policy is then fine-tuned against the reward model using PPO.
The authors found that optimizing their reward model results in better quality summaries than directly optimizing \rouge.

\subsection{Two-stage reranking approaches}

SimCLS \citep{liu-liu-2021-simcls} proposes formulating text generation as a reference-free quality estimation problem assisted by contrastive learning. The first stage decodes candidates with diverse beam search and a RoBERTa based model is used to rank them in the second stage.

SummaReRanker \citep{ravaut-etal-2022-summareranker} observes improved performance when training the generation and the reranking models on two non-overlapping halves of the fine-tuning data compared to training two models on the same data.

\citet{lee-etal-2021-discriminative} trains a discriminative reranker for neural machine translation that predicts the observed distribution of BLEU scores over the n-best list.

BRIO \citep{liu-etal-2022-brio} includes a two-stage reranking system that uses \seqtoseq generation models.
It is shown that the \seqtoseq reranker has better performance than encoder-only models in providing ranking scores. 

\subsection{Multi Task Learning with Sequence-level Loss}

\citet{edunov-etal-2018-classical} surveys a range of classical objective functions for structured prediction and apply them to \seqtoseq models.
Their experiments showed that combining sequence-level objectives with token-level objectives yields improved performance on translation and summarization datasets.

\citet{Sun2021AlleviatingEB} combines contrastive learning objective with negative log-likelihood to decrease the likelihood of the model generated ``silver" summaries meanwhile increasing the likelihood of the ``gold'' references.

BRIO \citep{liu-etal-2022-brio} demonstrates that multi task learning of sequence candidates with contrastive reranking and token-level generation has better performance compared to a two-stage reranking system.
The ranking order is determined by similarity to target using external metrics (\rouge, \bertscore). 
Models trained to rank by \rouge also perform well measured on \bertscore and vice versa.

\citet{DBLP:journals/corr/abs-2010-07447} extends label smoothing from classification tasks to semantic label smoothing for \seqtoseq learning.
Their technique adds sequence-level losses that smooth over well-formed relevant sequences that are similar to the target sequence semantically and on n-gram level.

\section{Conclusion}

We propose adding a third stage of sequence likelihood calibration (\slc) after the pretraining and fine-tuning stages for conditional language generation.
The calibration process decodes candidates from the fine-tuned model, and continues training to align their sequence likelihood according to their similarity to the target sequence in the model's latent space. 
A simple yet effective recipe for \slc is selecting the fine-tuned model’s checkpoint by perplexity, decoding candidates with beam search,  calibrating with \hinge loss and KL divergence regularization.
We are able to eliminate all decoding heuristics for calibrated models.
The benefits of calibration persist as models scale up in size.
Smaller calibrated models might outperform larger ones under the same inference compute budget.
By calibrating a \pegasusxxlarge model, we exceed or match state-of-the-art results on 8 datasets spanning abstractive summarization, generative question answering, question generation and data-to-text tasks.

\ifformatarxiv

\section*{Acknowledgement}
We thank David Grangier for early and engaging discussions, and Noah Fiedel for feedback on the paper.

\fi

\clearpage
\bibliography{anthology,custom}
\bibliographystyle{acl_natbib}

\newpage
\appendix

\section{Datasets Properties}
\label{appendix:dataset}

\subsection{Datasets and tasks}

\textbf{\cnn} \citep{hermann2015cnndm, see-etal-2017-get} summarization dataset contains 313k articles from the CNN and Daily Mail newspapers with bullet point summaries. The summaries are on average 3-4 sentences and relatively extractive.\footnote{\url{https://www.tensorflow.org/datasets/catalog/cnn_dailymail}}

\textbf{\xsum} \citep{narayan-etal-2018-dont} summarization dataset consists of 227k
BBC articles from 2010 to 2017 with a single sentence highly abstractive summary. Sometimes the summary contains information not present in the article.\footnote{\url{https://www.tensorflow.org/datasets/catalog/xsum}}

\textbf{\reddit} \citep{kim-etal-2019-reddittifu} summarization dataset contains 42k posts of informal stories from sub-reddit TIFU from 2013-Jan to 2018-Mar with author written summaries. The style and length of the summaries are very diverse.\footnote{\url{https://www.tensorflow.org/datasets/catalog/reddit_tifu}}

\textbf{\samsum} \citep{gliwa-etal-2019-samsum} summarization dataset contains 16k high-quality chat-dialogues and their summaries written by linguists.\footnote{\url{https://www.tensorflow.org/datasets/catalog/samsum}}

\textbf{\qg} \citep{nqg2017, du-etal-2017-learning} is the task of generating a question from a passage-answer pair extracted from the SQuAD dataset \citep{rajpurkar-etal-2016-squad}. In particular, we use the split of \citet{du-etal-2017-learning}, consisting of 75,722, 10,570, and 11,877 examples for training, validation, and testing, respectively.\footnote{\url{https://www.tensorflow.org/datasets/catalog/squad_question_generation}}

\textbf{\msmarco} \citep{msmarco} is a large scale dataset focused on machine reading comprehension and question answering. The original QA dataset consists of 1,010,916 queries. However, we work on the NLGEN data that is a subset of the QA data consisting of 182,669 queries, each with a well formed answer. The task is to generate a well formed answer to an input query and a set of answering passages.\footnote{\url{https://huggingface.co/datasets/ms_marco/viewer/v2.1}}

\textbf{\webnlg} \citep{gardent-etal-2017-webnlg} consists of 16,095 data inputs in the from of sets of RDF triples extracted from DBpedia. Each data point was verbalized by humans in more-than-one natural texts, leading to a total of 38,872 data-text pairs. \footnote{\url{https://www.tensorflow.org/datasets/catalog/gem\#gemweb_nlg_en}}

\textbf{\commongen} \citep{lin-etal-2020-commongen} introduces a tak of  generating a coherent sentence describing an input set of common concepts. The dataset consists of a total of 35,141 common concept sets, split into 32,651/993/1,497 training/validation/test sets. There are 67,389,	4,018 and 6,042 sentences in training, validation and test, respectively.\footnote{\url{https://www.tensorflow.org/datasets/catalog/gem\#gemcommon_gen_default_config}}

\begin{table}[tbh]
\caption{Statistics of datasets.}
\begin{center}

\small
\begin{tabular}{l c cc cc}
\hline
dataset & \# of examples  &  \multicolumn{2}{c}{avg words} &  \multicolumn{2}{c}{extractiveness} \\
& train/val/test  & input & target & coverage & density \\
\hline
\cnn & 287K / 13K / 11K &  698.60 &  49.53 &  87.8\% &  3.77 \\
\xsum &  203K / 11K / 11K &  383.17 &  21.74 &  63.9\% &  1.06 \\
\reddit &   34K / 4K / 4K &  396.15 &  21.02 &  68.4\% &  1.27 \\
\samsum &  14,732 / 818 / 819 &   97.23 &  21.00 &  68.0\% &  1.46 \\
\hline
\qg & 76K / 11K / 12K &  128.72 &  10.24 &  64.7\% &  1.63 \\
\msmarco & 152K / 12K / 12K &  588.50 &  14.07 &  97.5\% &  7.78 \\
\hline
\webnlg &  35K / 1667 / 1779 &   17.50 &  20.51 &  48.7\% &  1.3 \\
\commongen & 67K / 993 / 1497 &    3.27 &  10.10 &  22.0\% &  0.22 \\
\hline

\end{tabular}
\end{center}
\label{tab:datasets}
\end{table}

\section{Model Architecture}
\label{appendix:model_size}

Model sizes and their configurations are reported in \autoref{tab:appendix_architecture}.
\begin{table}[tbph!]
\caption{Model sizes.}
\begin{center}

\small
\begin{tabular}{l c cccc cc}
\hline
name & num layers & hidden size & num heads & MLP size & \multicolumn{2}{c}{\# num params}
\\
 & enc/dec  &  &  &  & excluding embs & \# total
\\
\hline
\pegasussmall & 8/8 & 512 & 8 & 1024 & 49M & 108M 
\\
\pegasusbase & 12/12 & 768 & 12 & 3072 & 198M & 272M
\\
\pegasuslarge & 16/16 & 1024 & 16 & 4096 & 470M & 568M 
\\
\pegasusxxlarge & 24/24 & 1024 & 16 & 16384 & 1913M & 2012M
\\

\hline

\end{tabular}
\end{center}
\label{tab:appendix_architecture}
\end{table}

\section{Ablation Study}
\label{appendix:ablation}

\slc methods for ablation study are reported in \autoref{tab:appendix_ablation}.
\begin{table}[tbh]
\caption{Experimental settings for ablation studies. }
\begin{center}

\small
\setlength{\tabcolsep}{4pt}
\begin{tabular}{ll cccccc cc}
\hline
\multicolumn{2}{l}{Ablation} &  \multicolumn{6}{c}{calibration} & \multicolumn{2}{c}{evaluation} \\
 & & decoding & sim fn & loss & regularization & ckpt & extra & decoding \\
\hline
 & fine-tuned & - & - & - & - & - & - & \bs 5
\\
\hline

\multicolumn{7}{l}{similarity function } \\

& ROUGE &
\bs 15 & ROUGE & \minrisk & \ce & \rouge & fix lr & \bs 5
\\
& decoder repr & 
\bs 15 & \simdec  & \minrisk & \ce & \rouge & fix lr & \bs 5
\\
& token emb & 
\bs 15 & \simtok & \minrisk & \ce & \rouge & fix lr & \bs 5
\\
\hline

\multicolumn{7}{l}{calibration loss } \\
& \hinge & 
\bs 15 & \simdec  & \hinge & \ce & \rouge & best lr, $\beta$& \bs 5
\\
& \maxmargin &
\bs 15 & \simdec  & \maxmargin & \ce & \rouge & best lr, $\beta$ & \bs 5
\\
& \listrank &
\bs 15 & \simdec  & \listrank & \ce & \rouge & best lr, $\beta$ & \bs 5
\\
& \minrisk & 
\bs 15 & \simdec  & \minrisk & \ce & \rouge & best lr, $\beta$ & \bs 5
\\
\hline

\multicolumn{7}{l}{regularization loss } \\
& none &
\bs 15 & \simdec  & \hinge & - & \rouge & fix lr, $\beta$ & \bs 5
\\
& \ce&
\bs 15 & \simdec  & \hinge & \ce & \rouge & fix lr, $\beta$ & \bs 5
\\
& \kl & 
\bs 15 & \simdec  & \hinge & \kl & \rouge & fix lr, $\beta$ & \bs 5
\\
\hline

\multicolumn{7}{l}{calibration decoding method} \\

& \bs & 
\bs 15 & \simdec  & \minrisk & \ce & \rouge & fix lr & \bs 5
\\
& \dbs &
\dbs 15 & \simdec  & \minrisk & \ce & \rouge & fix lr & \bs 5
\\
& \ns & 
\ns 15 & \simdec  & \minrisk & \ce & \rouge &fix lr  & \bs 5
\\
\hline

\multicolumn{7}{l}{calibration checkpoint selection} \\
& \rouge &
\bs 15 & \simdec  & \minrisk & \ce & \rouge & fix lr & \bs 5
\\
& \ppl &
\bs 15 & \simdec  & \minrisk & \ce & \ppl & fix lr & \bs 5
\\
\hline

\end{tabular}
\end{center}

\label{tab:appendix_ablation}
\end{table}

\newpage
\section{Decoding Methods}
\label{appendix:decoding_method}
\slc methods for decoding calibrated models are reported in \autoref{tab:appendix_calibrated_decoding}.
At evaluation time, models are decoded with 1, 2, 5, 10 and 20 candidates.
\rouge numbers in \autoref{fig:decoding} are reported in \autoref{tab:appendix_calibrated_decoding_numbers}.

\begin{table}[tbh!]
\caption{Experimental settings for calibrated models' decoding analysis. }
\begin{center}

\small
\setlength{\tabcolsep}{4pt}
\begin{tabular}{l cccccc cc}
\hline
name &  \multicolumn{6}{c}{calibration} & \multicolumn{2}{c}{evaluation} \\
& decoding & sim fn & loss & regularization & ckpt & extra & decoding & $\alpha$ \\
\hline
/ $\rightarrow$ beam &
 &  &  &  &  & & \bs 1-20 & \cmark
 \\
/ $\rightarrow$ nucleus &
 &  &  &  &  & & \ns 1-20 & \cmark
\\
beam $\rightarrow$ beam &
\bs 15 & \simdec & \minrisk & \ce & \rouge & fix lr & \bs 1-20 & \xmark
\\
beam $\rightarrow$ nucleus &
\bs 15 & \simdec & \minrisk & \ce & \rouge & fix lr & \ns 1-20& \xmark
\\
nucleus $\rightarrow$ beam &
\ns 15 & \simdec & \minrisk & \ce & \rouge & fix lr & \bs 1-20 & \xmark
\\
nucleus $\rightarrow$ nucleus &
\ns 15 & \simdec & \minrisk & \ce & \rouge & fix lr & \ns 1-20& \xmark
\\
\hline

\end{tabular}
\end{center}

\label{tab:appendix_calibrated_decoding}
\end{table}

\begin{table}[tbh]
\caption{ROUGE (\rouges) numbers of the decoding curves.}
\begin{center}

\small
\setlength{\tabcolsep}{4pt}
\begin{tabular}{ll cccc}
\hline
\slc $\rightarrow$ decoding & num  & {\cnn} & {\xsum} & {\reddit} & {\samsum} \\
 & decodes & \rouges & \rouges & \rouges & \rouges \\ 
\\
\hline

/ $\rightarrow$ \bs
& 1 & 45.11/21.15/42.34 & 46.18/22.84/38.07 & 27.78/8.40/22.18 & 52.86/27.89/43.85 
\\
& 2 & 44.54/21.62/41.73 & 46.94/23.87/38.89 & 27.75/8.98/22.37 & 53.42/29.11/44.56 
\\
& 5 & 44.78/21.99/41.93 & 47.26/24.38/39.24 & 26.88/9.09/21.95 & 53.47/29.25/44.53 
\\
& 10 & 44.58/21.86/41.71 & 47.29/24.60/39.41 & 25.51/8.78/21.04 & 53.70/29.22/44.63 
\\
& 20 & 44.33/21.64/41.43 & 47.13/24.62/39.36 & 24.10/8.32/20.06 & 53.74/29.21/44.64 
\\

/ $\rightarrow$ \ns
& 1 & 44.09/19.88/41.24 & 43.76/20.42/35.51 & 25.33/6.84/19.78 & 50.51/24.56/40.85 
\\
& 2 & 44.31/20.36/41.50 & 45.03/21.80/36.96 & 24.82/6.95/19.72 & 52.17/26.91/43.02 
\\
& 5 & 44.43/20.81/41.67 & 45.61/22.63/37.83 & 23.80/6.84/19.43 & 51.50/26.70/43.02 
\\
& 10 & 44.28/20.94/41.54 & 46.06/23.22/38.37 & 23.45/6.98/19.44 & 50.69/26.28/42.70 
\\
& 20 & 44.25/21.13/41.53 & 46.06/23.57/38.62 & 21.87/6.81/18.39 & 50.53/26.58/42.72 
\\

\bs $\rightarrow$ \bs
& 1 & 45.72/20.87/42.89 & 46.71/23.16/38.65 & 30.00/9.20/23.82 & 54.24/28.67/44.59 
\\
& 2 & 46.46/21.96/43.58 & 47.46/24.17/39.47 & 30.24/9.56/24.11 & 54.68/29.71/45.02 
\\
& 5 & 46.72/22.55/43.87 & 47.88/24.79/40.05 & 30.25/9.80/24.26 & 54.78/29.75/45.27 
\\
& 10 & 46.81/22.67/43.95 & 47.83/24.82/40.06 & 30.31/9.89/24.39 & 54.63/30.01/45.20 
\\
& 20 & 46.90/22.83/44.04 & 47.83/24.86/40.07 & 30.02/9.80/24.29 & 54.74/29.98/45.15 
\\

\bs $\rightarrow$ \ns
& 1 & 44.83/19.59/41.92 & 44.73/20.99/36.52 & 28.19/7.89/22.05 & 52.26/26.19/42.07 
\\
& 2 & 45.16/20.01/42.28  & 45.55/21.92/37.56 & 28.66/8.22/22.58 & 53.15/27.61/43.45 
\\
& 5 & 45.35/20.34/42.49 & 46.15/22.87/38.45 & 28.83/8.62/23.06 & 53.50/27.80/44.18 
\\
& 10 & 45.46/20.51/42.59 & 46.39/23.33/38.85 & 28.90/9.10/23.47 & 53.99/28.71/44.89 
\\
& 20 & 45.46/20.63/42.63 & 46.53/23.67/39.07 & 28.60/9.01/23.39 & 54.22/28.68/45.19 
\\

\ns $\rightarrow$ \bs
& 1 & 45.66/20.93/42.77 & 46.50/22.93/37.97 & 30.57/9.45/23.68 & 53.81/28.71/44.23 
\\
& 2 & 46.19/21.91/43.29 & 47.29/23.93/38.90 & 30.94/9.82/24.06 & 53.99/29.25/44.30 
\\
& 5 & 46.47/22.50/43.56 & 47.74/24.43/39.36 & 31.10/10.00/24.22 & 54.29/29.49/44.62 
\\
& 10 & 46.39/22.57/43.52 & 47.78/24.52/39.41 & 31.02/10.00/24.22 & 54.25/29.54/44.70 
\\
& 20 & 46.34/22.63/43.48 & 47.83/24.63/39.49 & 31.11/10.09/24.29 & 54.17/29.46/44.59 
\\

\ns $\rightarrow$ \ns
& 1 & 44.68/19.69/41.75 & 44.35/20.69/35.80 & 29.85/8.68/22.94 & 52.55/26.98/42.63 
\\
& 2 & 45.14/20.24/42.20 & 45.50/21.94/37.13 & 30.31/9.22/23.45 & 52.97/27.33/43.00 
\\
& 5 & 45.58/20.81/42.65 & 46.43/22.93/38.19 & 30.46/9.44/23.81 & 54.10/28.86/44.77 
\\
& 10 & 45.73/21.05/42.82 & 46.91/23.65/38.90 & 30.69/9.58/24.11 & 54.02/28.82/44.70 
\\
& 20 & 45.78/21.26/42.88 & 47.19/24.00/39.14 & 31.04/9.89/24.44 & 53.78/29.02/44.66 
\\

\hline

\end{tabular}
\end{center}

\label{tab:appendix_calibrated_decoding_numbers}
\end{table}

\newpage
\section{Length Normalization}
\label{appendix:brevity_penalty}
Experimental settings for length normalization analysis is reported in \autoref{tab:appendix_brevity}.
Brevity penalty $\alpha$ is chosen as the best value for fine-tuned models' ROUGE performance on validation dataset or disabled.
\begin{table}[tbh!]
\caption{Experimental settings for length normalization study. }
\begin{center}

\small
\setlength{\tabcolsep}{6pt}
\begin{tabular}{ll cccccc cc}
\hline
\slc & $\alpha$ &  \multicolumn{6}{c}{calibration} & \multicolumn{2}{c}{evaluation} \\
& & decoding & sim fn & loss & regularization & ckpt & extra & decoding & $\alpha$ \\
\hline
\xmark & \xmark & 
 &  &  &  &  & & \bs 5 & \xmark
\\
\xmark & \cmark &
 &  &  &  &  & & \bs 5 & \cmark
\\
\cmark & \xmark &
\bs 15 & \simdec & best & \ce & \rouge & best lr, $\beta$ & \bs 5 & \xmark
\\
\cmark & \cmark &
\bs 15 & \simdec & best & \ce & \rouge & best lr, $\beta$ & \bs 5 & \cmark
\\
\hline

\end{tabular}
\end{center}

\label{tab:appendix_brevity}
\end{table}

\section{Model Flops Estimation}
\label{appendix:flops}
We extends formulations in Table 1 of \citet{scaling_law} to estimate FLOPs of our transformer encoder decoder models following the formula:
\begin{equation}
\begin{split}
total\_C & = C_{enc} \times n_{enc-ctx} + C_{dec} \times n_{dec-ctx} \times m \\
C_{enc} & = 2N_{enc} + 2n_{enc-layer}n_{enc-ctx}d_{enc-attn} \\
C_{dec} & = 2N_{dec} + n_{dec-layer}n_{dec-ctx}d_{dec-attn} \\
\end{split}
\end{equation}
where $m$ is the number of decoder candidates, other notations can be referenced in Table 1 of \citet{scaling_law}. Because of upper triangle attention masking, the effective decoder attention context length is half of sequence lengths instead of full sequence lengths as in the encoder. Extra computation incurred by different decoding methods are omitted as they are much smaller.

\section{Scaling}
\label{appendix:scaling}
\slc method for scaling curves are reported in \autoref{tab:appendix_scaling}.
At evaluation time, models are decoded with 1, 2, 5, 10, and maybe 15, 20 candidates.
\rouge numbers in \autoref{fig:scaling} are reported in \autoref{tab:appendix_scaling_numbers}.

\begin{table}[tbh!]
\caption{Experimental settings for scaling. }
\begin{center}

\small
\setlength{\tabcolsep}{6pt}
\begin{tabular}{l cccccc cc}
\hline
model &  \multicolumn{6}{c}{calibration} & \multicolumn{2}{c}{evaluation} \\
& decoding & sim fn & loss & regularization & ckpt & extra & decoding \\
\hline
fine-tuned &
&  &  &  &  & & \bs 1-20 
\\
calibrated &
\bs 15 & \simdec & \minrisk & \ce & \rouge & best lr & \bs 1-20 
\\
\hline

\end{tabular}
\end{center}

\label{tab:appendix_scaling}
\end{table}

\newpage
\begin{table}[tbh]
\caption{ROUGE (\rouges) numbers of the scaling curve.}
\begin{center}

\small
\setlength{\tabcolsep}{4pt}
\begin{tabular}{ll cccc}
\hline
size & decodes  & {\cnn} & {\xsum} & {\reddit} & {\samsum} \\
 &  & \rouges & \rouges & \rouges & \rouges \\ 
\\
\hline
\multicolumn{6}{c}{fine-tuned } \\

\smallmodelsize & 1 & 43.21/19.99/40.53 & 40.91/17.80/32.98 & 25.37/6.99/20.19 & 49.78/24.45/40.67
\\
& 2 & 42.77/20.40/39.94 & 41.55/18.78/33.75 & 25.22/7.53/20.34 & 50.52/25.37/41.80
\\
& 5 & 42.92/20.45/39.96 & 41.87/19.44/34.28 & 24.41/7.61/20.00 & 50.52/25.92/42.00
\\
& 10 & 42.78/20.32/39.75 & 41.85/19.57/34.38 & 23.04/7.43/19.04 & 50.41/25.84/41.81
\\
& 15 & - & 41.79/19.59/34.31 & - & 50.46/25.89/41.77
\\
& 20 & - & 41.65/19.56/34.25 & - & 50.50/26.00/41.45
\\

\basemodelsize & 1 & 44.59/20.96/41.93 & 44.51/21.34/36.47 & 27.32/8.06/21.56 & 51.77/26.44/42.38
\\
& 2 & 44.06/21.44/41.33 & 45.24/22.22/37.15 & 27.36/8.49/21.89 & 52.35/27.40/43.27
\\
& 5 & 44.08/21.54/41.27 & 45.65/22.83/37.71 & 26.61/8.78/21.67 & 52.48/27.72/43.70
\\
& 10 & 43.84/21.30/40.96 & 45.61/22.93/37.70 & 25.80/8.37/20.85 & 52.40/27.64/43.67
\\
& 15 & - & 45.55/22.94/37.71 & - & 52.35/27.69/43.67
\\
& 20 & - & 45.54/22.99/37.71 & - & 52.38/27.68/43.68
\\

\largemodelsize & 1 & 45.34/21.47/42.60 & 46.27/23.02/38.12 & 27.79/8.42/22.18 & 53.05/27.96/43.66
\\
& 2 & 44.93/21.83/42.15 & 46.99/23.90/38.89 & 27.76/8.99/22.36 & 53.73/29.07/44.75
\\
& 5 & 44.78/21.98/41.92 & 47.26/24.37/39.23 & 26.85/9.09/21.94 & 53.94/29.01/44.53
\\
& 10 & 44.59/21.86/41.71 & 47.27/24.59/39.40 & 25.97/8.74/20.99 & 53.67/29.31/44.62
\\
& 15 & - & 47.20/24.63/39.41 & - & 53.71/29.22/44.63
\\
& 20 & - & 47.15/24.62/39.37 & - & 53.68/29.16/44.61
\\

\xxlargemodelsize & 1 & 45.52/21.70/42.73 & 47.89/24.54/39.67 & 28.82/9.29/23.13 & 53.40/28.01/43.82
\\
& 2 & 45.37/21.95/42.54 & 48.66/25.61/40.55 & 28.60/9.60/23.12 & 53.89/29.47/44.88
\\
& 5 & 45.40/22.09/42.56 & 48.94/26.18/40.91 & 27.86/9.87/22.84 & 53.98/29.08/44.62
\\
& 10 & 45.29/21.82/42.44 & 48.91/26.08/40.84 & 27.52/9.01/21.86 & 53.95/29.61/44.61
\\
& 15 & - & 48.96/26.12/40.78 & - & 53.92/29.61/44.63
\\
& 20 & - & 48.75/26.20/40.83 & - & 53.86/29.57/44.59
\\

\hline
\multicolumn{6}{c}{calibrated } \\

\smallmodelsize & 1 & 44.31/20.82/41.65 & 41.41/17.95/33.15 & 27.15/7.57/21.48 & 49.85/24.62/40.33
\\
& 2 & 44.91/21.76/42.13 & 42.27/18.99/34.11 & 27.34/8.00/21.70 & 50.89/25.71/41.82
\\
& 5 & 45.12/22.10/42.25 & 42.88/19.84/34.89 & 27.49/8.43/22.02 & 51.53/26.58/42.34
\\
& 10 & 45.15/22.20/42.22 & 43.01/20.13/35.11 & 27.32/8.42/21.92 & 52.08/26.67/42.17
\\
& 15 & - & 43.13/20.15/35.16 & - & 52.04/26.66/42.10
\\
& 20 & - & 43.14/20.19/35.21 & - & 51.90/26.71/42.16
\\

\basemodelsize & 1 & 45.26/21.16/42.57 & 44.54/21.18/36.27 & 27.81/8.10/21.80 & 52.12/26.48/42.40
\\
& 2 & 45.97/22.25/43.21 & 45.47/22.12/37.18 & 28.38/8.68/22.38 & 53.29/28.24/43.92
\\
& 5 & 46.18/22.78/43.41 & 46.04/22.90/37.86 & 28.51/9.03/22.79 & 53.79/28.75/44.15
\\
& 10 & 46.26/22.88/43.47 & 46.21/22.99/38.01 & 28.35/9.07/22.78 & 54.06/28.86/44.49
\\
& 15 & - & 46.29/23.07/38.05 & - & 54.03/28.85/44.41
\\
& 20 & - & 46.28/23.09/38.03 & - & 53.99/28.90/44.39
\\

\largemodelsize & 1 & 45.55/20.85/42.76 & 46.42/22.93/38.12 & 29.29/9.10/23.26 & 53.31/28.43/44.18
\\
& 2 & 46.30/21.92/43.43 & 47.29/23.95/39.02 & 29.80/9.59/23.75 & 54.14/29.29/44.47
\\
& 5 & 46.55/22.48/43.68 & 47.88/24.62/39.62 & 29.83/9.84/23.91 & 54.61/29.95/45.10
\\
& 10 & 46.63/22.58/43.78 & 47.93/24.74/39.76 & 29.87/9.95/24.03 & 54.89/30.05/45.18
\\
& 15 & - & 48.05/24.80/39.83 & - & 54.88/30.27/45.34
\\
& 20 & - & 48.06/24.85/39.86 & - & 54.87/30.31/45.39
\\

\xxlargemodelsize & 1 & 46.29/21.92/43.47 & 48.11/24.59/39.68 & 30.20/9.86/24.15 & 54.71/29.45/45.03
\\
& 2 & 46.84/22.93/43.95 & 49.04/25.55/40.46 & 30.59/10.38/24.50 & 55.17/30.68/46.09
\\
& 5 & 47.08/23.45/44.19 & 49.56/26.31/41.08 & 30.65/10.70/24.76 & 55.46/30.71/46.11
\\
& 10 & 47.08/23.57/44.19 & 49.79/26.56/41.32 & 30.75/10.79/24.91 & 55.47/30.60/46.00
\\
& 15 & - & 49.79/26.55/41.35 & - & 55.41/30.63/46.15
\\
& 20 & - & 49.76/26.54/41.30 & - & 55.38/30.65/46.14
\\

\hline

\end{tabular}
\end{center}

\label{tab:appendix_scaling_numbers}
\end{table}

\newpage
\section{Final Results}
\label{appendix:final}

\slc method for final results is reported in \autoref{tab:appendix_final}.
We choose the \slc best based on \autoref{sec:ablation}.
There are in total 3 hyper-parameters: learning rate $lr$ (\hyperref[alg:calibration]{Algorithm \ref*{alg:calibration}}), ranking constant $\beta$ (\autoref{eq:seq_loss}), and regularization strength $\lambda$ (\autoref{eq:reg_loss}).
We fix two of the them: $\beta$ is set to 10, and $lr * \lambda$ is set to $1e-5$.
Best learning rate $lr$ is determined with hyper-parameter tuning on validation set and reported in \autoref{tab:appendix_final_lr}.

\begin{table}[tbh!]
\caption{Experimental settings for length normalization study. }
\label{tab:appendix_final}
\begin{center}

\small
\setlength{\tabcolsep}{6pt}
\begin{tabular}{l cccccc cc}
\hline
model &  \multicolumn{6}{c}{calibration} & \multicolumn{2}{c}{evaluation} \\
& decoding & sim fn & loss & regularization & ckpt & extra & decoding \\
\hline
fine-tuned &
&  &  &  &  & & \bs 5 
\\
calibrated &
\bs 15 & \simdec & \hinge & \kl & \ppl & best lr & \bs 10
\\
\hline

\end{tabular}
\end{center}

\end{table}

\begin{table}[tbh]
\caption{Learning rate of final results.}
\label{tab:appendix_final_lr}
\begin{center}
\small
\begin{tabular}{l cccc}
\hline 
 & {\cnn} & {\xsum} & {\reddit} & {\samsum} \\
lr & $10^{-5}$ & $10^{-5}$ & $10^{-5}$ & $10^{-6}$\\
\hline
 & {\msmarco} & {\qg} & {\webnlg} & {\commongen} \\
lr & $3\times10^{-6}$ & $10^{-5}$ & $10^{-6}$ & $10^{-5}$\\
\hline
\end{tabular}
\end{center}
\end{table}

\end{document}